\documentclass{article}

\usepackage[preprint]{corl_2026} 
\usepackage{amssymb}            
\usepackage{mathtools}          
\usepackage{mathrsfs}           
\usepackage{graphicx}           
\usepackage{subcaption}         
\usepackage[space]{grffile}     
\usepackage{url}                
\usepackage{lipsum}             
\usepackage{amsmath}
\usepackage{algorithm}
\usepackage{algpseudocode}

\title{Curriculum Generation under Structured Parametric Environments for Robust Navigation Policies}

\author{
  Prishita Ray\thanks{This research was conducted while the author was affiliated with Cornell University.}\\
  Cornell University \\
  pr376@cornell.edu \\
}

\begin{document}
\maketitle


\begin{abstract}
Robust navigation policies for autonomous agents must generalize across continuously varying environmental conditions such as turn rates, obstacles, friction, pits, and slopes. Curriculum generation provides a principled mechanism for improving generalization by progressively adapting training environments, but designing such curricula in a sample-efficient and automated manner remains challenging.
This paper proposes a reparameterized curriculum generation framework for structured continuous environment parameters using unidirectional gradient-based optimization. To improve robustness in multimodal observation spaces consisting of image-based and scalar inputs, a distribution-shift regularization objective is incorporated to encourage the learning of finer-grained latent representations. The proposed method is evaluated across two continuous-control OpenAI Gym environments: a 2D obstacle-based Car Racing variant and Bipedal Walker variant, where coupled environment parameters jointly influence policy performance. Across five random seeds, our method consistently outperforms vanilla policy training, random parameter sampling, manual curricula, frontier-based methods, Self-Paced Reinforcement Learning (SPRL), Absolute Learning Progress with Gaussian Mixture Models (ALP-GMM), and reverse curriculum learning baselines. Ablation studies further demonstrate the effectiveness of the reparameterized curriculum mechanism across both environments, while highlighting environment-dependent benefits of the auxiliary regularization objective.
\end{abstract}



\section{Introduction}

Model-free reinforcement learning (RL) is widely used for learning navigation and locomotion policies in simulated robotic environments. However, training on fixed environment distributions often limits robustness to deployment variation. Curriculum learning addresses this challenge by adaptively selecting training environments \citep{bengio2009curriculum,sullivan2025syllabus}. Existing curriculum methods commonly rely on black-box optimization \citep{BO2023} or adversarial/minimax formulations \citep{REPAIRED, CLUTR}, which can be computationally expensive or unstable in continuous parameterized environments.

This paper proposes a reparameterized curriculum generation framework for low-dimensional continuous environment parameter spaces. Our contributions are:
\begin{itemize}
    \item A gradient-based curriculum adaptation method over coupled continuous environment parameters.
    \item A distribution-shift regularization objective for more stable curriculum evolution in image-based settings.
    \item Evaluation on modified Car Racing and Bipedal Walker environments with continuously varying difficulty parameters.
    \item Comparisons against vanilla RL, random sampling, manual curricula, SPRL, ALP-GMM, frontier-based, and reverse curriculum baselines across five seeds.
\end{itemize}


\section{Related Work}
\label{sec:related_work}

Curriculum learning and adaptive environment generation have been widely studied for improving robustness and generalization in RL \citep{wang2019autonomous, song2022robust, narvekar2020curriculum, portelas2020automatic}.

\textbf{Unsupervised Environment Design (UED)} methods generate adversarial or progressively difficult tasks using teacher policies or environment generators \citep{matiisen2019teacher, wang2019poet, parkerholder2022accelerating, PAIRED, REPAIRED, DRAGEN, CLUTR}. These methods have been applied across navigation, procedural generation, and robotics domains \citep{cobbe2020leveraging, anzalone2021reinforced, qiao2018automatically, ryu2025curricullm}. While effective, many rely on black-box or population-based optimization that can become computationally expensive in continuous parameterized environments. 

\textbf{Robust and minimax RL} approaches formulate training as a game against adversarial perturbations \citep{pinto2017robust, vinitsky2020robust, moos2022robust}. Bayesian optimization has also been explored for curriculum adaptation \citep{BO2023}, though such approaches remain computationally intensive and non-differentiable.

\textbf{Automatic curriculum learning} methods such as SPRL \citep{klink2020self}, CURROT \citep{klink2022curriculum}, ALP-GMM \citep{portelas2020teacher}, PLR \citep{jiang2021prioritized} and ACuTE \citep{shukla2022acute} adapt continuously parameterized environments using competence- or progress-based sampling strategies. However, these methods rely on stochastic task selection rather than direct gradient-based updates over coupled environment parameters.

\textbf{Domain randomization} improves robustness by sampling diverse environment parameters \citep{tobin2017domain}, but static sampling does not adapt to learner competence.

Our method performs gradient-based curriculum adaptation directly over continuous environment parameter distributions, enabling efficient and controlled curriculum evolution in structured parametric environments.


\section{Problem Formulation}
\label{sec:problemformulation}

\subsection{Car Racing Obstacles Environment}

This paper uses a modified OpenAI Gym Car Racing environment parameterized by road curvature $\kappa$ and obstacle probability $p$, where higher values correspond to more difficult tracks. An environment is represented as $e=(\kappa,p)$.

\begin{figure}[t]
    \centering
    \begin{subfigure}{0.5\linewidth}
        \centering
        \includegraphics[width=\linewidth]{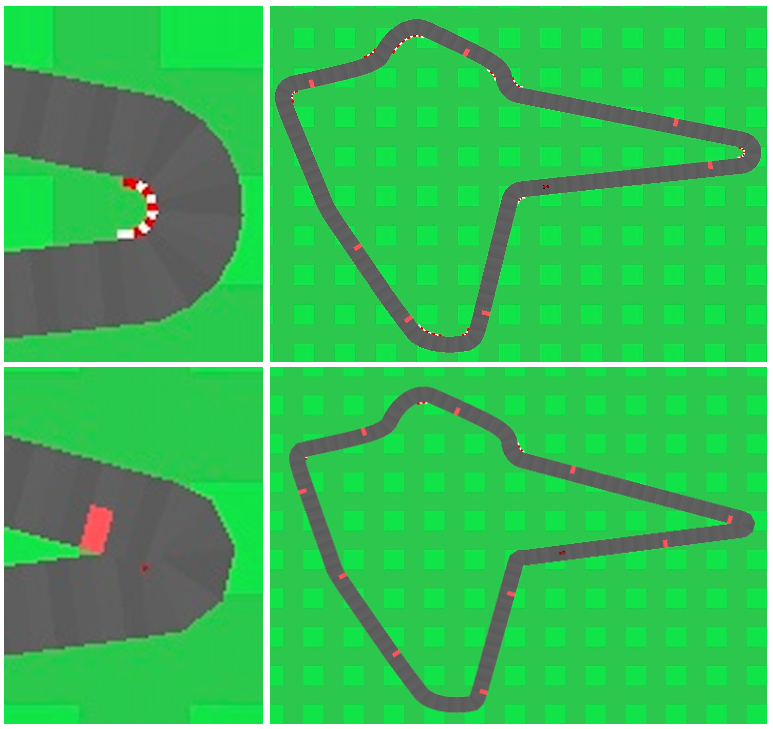}
        \caption{Top: [0.31, 0.05], Bottom: [0.71, 0.13]}
        \label{fig1a}
    \end{subfigure}
    \hfill
    \begin{subfigure}{0.48\linewidth}
        \centering
        \includegraphics[width=\linewidth,height=0.975\linewidth]{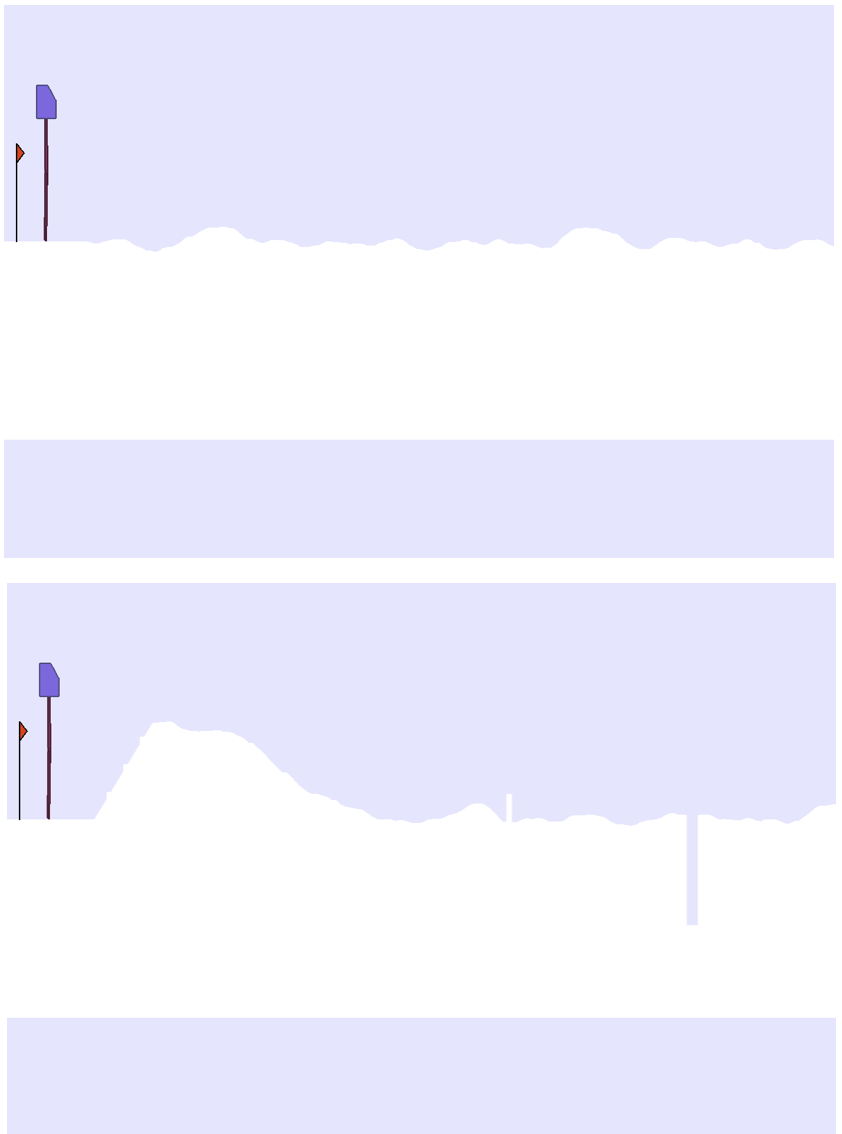}
        \caption{Top: [2.5, 0.0, 0.0], Bottom: [3.0, 0.2, 0.01]}
        \label{fig1b}
    \end{subfigure}

    \caption{Examples of Car Racing and Bipedal Walker environments under varying parameter settings.}
    \label{fig1}
\end{figure}

Observations consist of RGB bird's-eye-view images together with $(\kappa,p)$, while actions correspond to steering, acceleration, and braking. Rewards encourage track completion while penalizing delays and obstacle collisions, including a $-50$ penalty per obstacle collision:
\begin{equation}
R_\mathrm{t} = \frac{1000}{M}|T| - 50|O| - 0.1N
\label{eq1}
\end{equation}
where $M$ denotes the total number of road tiles, $T$ the traversed tiles, $O$ the number of obstacle collisions, and $N$ the number of delays. Figure \ref{fig1}a shows example environments for varying $\kappa$ and $p$.

\subsection{Bipedal Walker Environment}

Additionally the method is evaluated on a modified OpenAI Gym Bipedal Walker environment parameterized by terrain friction $f$, pit frequency $\rho$, and terrain slope magnitude $s$, represented as $e=(f,\rho,s)$.

Observations consist of proprioceptive state features together with $(f,\rho,s)$, while actions correspond to continuous motor torques. Rewards encourage stable forward locomotion while penalizing unstable posture, excessive torque usage and falls:
\begin{equation}
R_t = \Delta \left(130 \frac{p_t}{S}- 5|h_t|\right)- \lambda_u \sum_{i=1}^{4}|\tau_{t,i}| - 100 \cdot \mathbb{I}_{\mathrm{fall}}
\label{eq2}
\end{equation}
where $p_t$ denotes hull position, $h_t$ the hull angle, and $\tau_{t,i}$ the motor torque applied at joint $i$ and $\lambda_u = 0.028$. Figure \ref{fig1}b shows example terrain configurations.

Given a set of environments $\mathbf{E}$, the objective is to learn a policy $\pi:\mathcal{O}\rightarrow\mathcal{A}$ maximizing:
\begin{equation*}
\mathbb{E}_{e \sim \text{Unif}(\mathbf{E})}
\left[
\sum_{t=1}^{T}R_t
\right].
\end{equation*}

\section{Methodology}
\label{sec:methodology}

\begin{figure*}[t]
    \centering
    \includegraphics[scale=0.37]{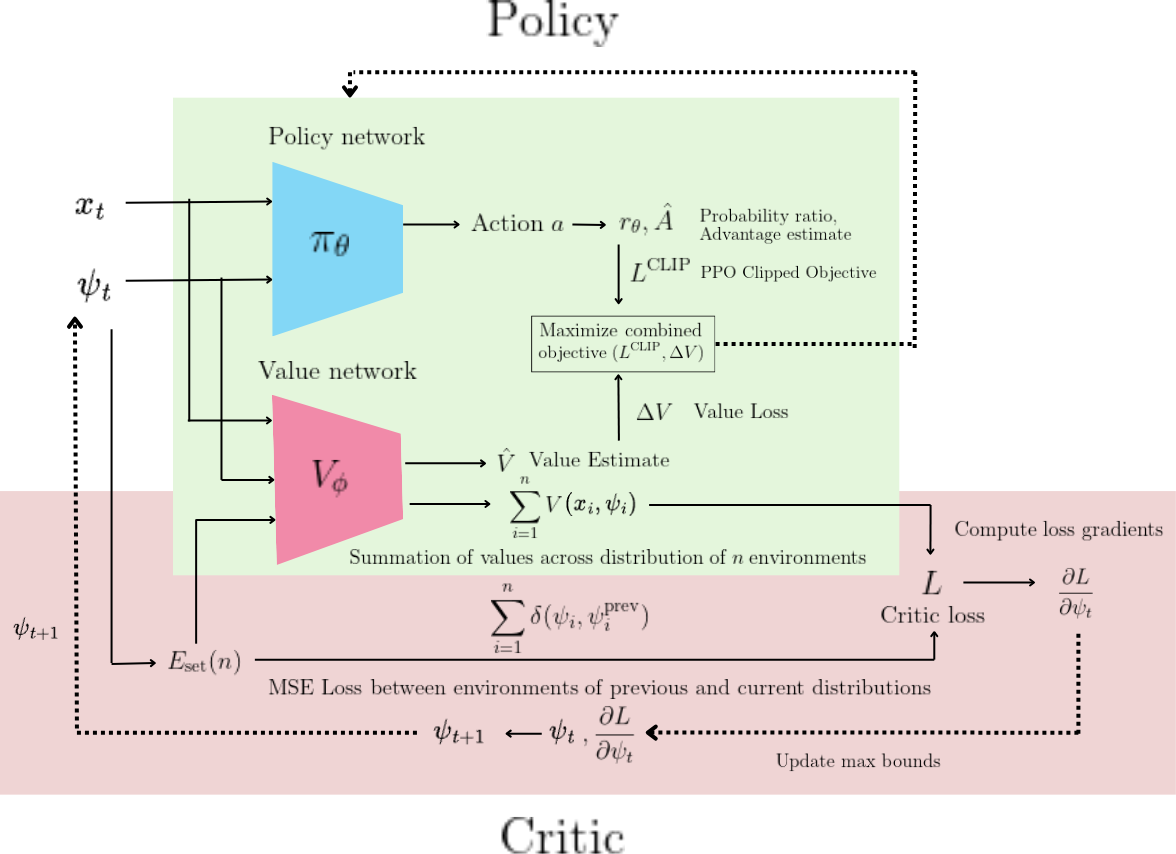}
    \caption{Overview of the Reparam framework. The policy network (green) is optimized using PPO, while the critic (red) estimates policy value across sampled environments and updates curriculum bounds over environment parameters $\psi_t$.}
    \label{fig2}
\end{figure*}

\subsection{Reparameterization Curriculum}

Two modules are used in the Reparameterization Curriculum (Reparam): the policy and the critic (see Fig. \ref{fig2} and Algorithm \ref{algo1}). The policy network parameterized by $\theta$ learns control policies conditioned on environment observations $x$ and environment parameters $\psi$, while the critic uses the value function network parameterized by $\phi$ adapts curriculum difficulty through gradient-based updates over continuous environment parameter bounds.

For Car Racing, the policy uses a multi-input architecture where RGB observations are processed using a CNN and environment parameters $\psi=(\kappa,p)$ are encoded using an MLP before feature fusion. For Bipedal Walker, proprioceptive states are concatenated with $\psi=(f,\rho,s)$ and processed using an MLP. PPO is used to optimize the policy using clipped policy and value objectives.

\subsection{Curriculum Critic}

The critic adapts the curriculum by estimating policy value across environments sampled within the current curriculum bounds. Let $\psi_{\mathrm{min}}$ and $\psi_t$ denote the lower and current upper curriculum bounds respectively. Environments are sampled as:
\begin{equation*}
\psi_i \sim U[\psi_{\mathrm{min}}, \psi_t].
\end{equation*}

The critic objective is computed over $n$ sampled environments:
\begin{equation*}
L =
\frac{1}{n}
\sum_{i=1}^{n}
V_\phi(x_i,\psi_i),
\end{equation*}
where $V_\phi$ denotes the PPO value network.

To enable gradient-based curriculum updates, environment sampling is reparameterized as:
\begin{equation*}
\psi_i =
\psi_{\mathrm{min}}
+
z_i \odot
(\psi_t-\psi_{\mathrm{min}}),
\qquad
z_i \sim U[0,1].
\end{equation*}

For the $j^{\mathrm{th}}$ parameter:
\begin{equation*}
\psi_i[j]
=
\psi_{\mathrm{min}}[j]
+
z_i[j]
(
\psi_t[j]-\psi_{\mathrm{min}}[j]
),
\end{equation*}

giving:
\begin{equation*}
\frac{\partial \psi_i[j]}
{\partial \psi_t[j]}
=
z_i[j].
\end{equation*}

Optionally, a distribution-shift regularization term is included between the current and previous curriculum distributions:
\begin{equation*}
\delta(\psi_i,\psi_i^{\mathrm{prev}})
=
\left\|
\psi_i-\psi_i^{\mathrm{prev}}
\right\|^2.
\end{equation*}

The overall critic objective becomes:
\begin{equation*}
\mathcal{L}
=
\frac{1}{n}
\sum_i
\left(
V_\phi(x_i,\psi_i)
-
\beta
\delta(\psi_i,\psi_i^{\mathrm{prev}})
\right).
\end{equation*}

Negative gradients indicate that increasing curriculum bounds decreases estimated policy value, corresponding to more difficult environments. Curriculum learning rates are annealed during training:
\begin{equation}
lr_{\psi[j],t}^{\mathrm{anneal}}
=
\frac{lr_{\psi[j]}}{\alpha_t}.
\end{equation}

Additionally curriculum updates are normalized using:
\begin{equation}
\eta_{\psi[j],t}
=
\frac{
\psi_t[j]-\psi_{\mathrm{min}}[j]
}{
\psi_{\mathrm{max}}[j]-\psi_{\mathrm{min}}[j]
}.
\end{equation}

The curriculum upper bounds are updated as:
\begin{equation*}
\psi_{t+1}[j] =
\begin{cases}
\psi_t[j]
-
lr_{\psi[j],t}^{\mathrm{anneal}}
\left(
\eta_{\psi[j],t}
\dfrac{\partial \mathcal{L}}
{\partial \psi_t[j]}
\right),
&
\text{if }
\dfrac{\partial \mathcal{L}}
{\partial \psi_t[j]}
< 0
\\
\psi_t[j],
&
\text{otherwise}.
\end{cases}
\end{equation*}

\begin{algorithm}[t]
\caption{Reparam Curriculum Generator}
\begin{algorithmic}[1]
\State Initialize policy parameters $\theta$, critic parameters $\phi$
\Repeat
    \State Collect trajectories using PPO policy $\pi_{\theta}$
    \State Update PPO policy and value networks
    \State Sample environments $\psi_i \sim U[\psi_{\min},\psi_t]$
    \State Compute critic objective:
    \[
    \mathcal{L}
    =
    \frac{1}{n}
    \sum_i
    \left(
    V_\phi(x_i,\psi_i)
    -
    \beta \delta(\psi_i,\psi_i^{\mathrm{prev}})
    \right)
    \]
    \State Compute gradients $\frac{\partial \mathcal{L}}{\partial \psi_t[j]}$
    \State Update curriculum upper bounds $\psi_t$
\Until{convergence}
\end{algorithmic}
\label{algo1}
\end{algorithm}

\section{Results and Discussion}
\label{sec:result}

\subsection{Metrics}
\label{sec:metrics}
The following metrics are used to evaluate performance across environments:

\textbf{Car Racing}
\begin{enumerate}
    \item Training time
    \item Mean and standard deviation of episodic rewards
    \item Number of obstacle collisions
    \item Collision ratio relative to the total number of encountered obstacles
    \item Ratio of time spent on grass to on-road traversal time
    \item Number of track tiles visited
\end{enumerate}

\textbf{Bipedal Walker}
\begin{enumerate}
    \item Training time
    \item Mean and standard deviation of episodic rewards
    \item Pit-fall ratio relative to the total number of encountered pits
    \item Slope-failure termination ratio
    \item Slip-failure termination ratio
    \item Average number of walking steps completed
\end{enumerate}

The interquartile mean scores for each of these metrics are reported in the Appendix.

\subsection{Baselines and Ablations}
\label{sec:baselines}

Reparam is compared against the following baselines on the test environments $\mathbf{E}_{\mathrm{test}}$:

\textbf{Vanilla Policy:}
Training is performed only on the default minimum-difficulty environment parameters $\psi_{\mathrm{min}}$ without curriculum adaptation.

\textbf{Random Sampling:}
Training environments are uniformly sampled from the full parameter range
$\psi_i \sim U[\psi_{\mathrm{min}},\psi_{\mathrm{max}}]$
without curriculum growth.

\textbf{Manual Curriculum:}
The curriculum upper bounds $\psi_t$ increase linearly from $\psi_{\mathrm{min}}$ to $\psi_{\mathrm{max}}$ throughout training.

\textbf{SPRL:}
Self-Paced Reinforcement Learning (SPRL) \citep{klink2020self} maintains a Gaussian curriculum distribution $\psi_i \sim \mathcal{N}(\mu_t,\Sigma_t)$ where the distribution parameters $(\mu_t,\Sigma_t)$ are adapted using policy competence estimates and progressively shifted toward more difficult environments.

\textbf{ALP-GMM:}
Absolute Learning Progress with Gaussian Mixture Models (ALP-GMM) \citep{portelas2020teacher} maintains a Gaussian mixture curriculum distribution
$\psi_i \sim \sum_{k=1}^{K} w_k \mathcal{N}(\mu_k,\Sigma_k)$ where environments are sampled from regions associated with high absolute learning progress.

\textbf{Frontier Curriculum:}
A gradient-based curriculum variant where environments are sampled only from a local frontier region $\psi_i \in (\psi_t-\Delta,\psi_t)$ instead of the full running curriculum range $\psi_i \in (\psi_{\mathrm{min}},\psi_t)$.

\textbf{Reparam-R Curriculum:}
A reverse curriculum variant that adapts lower curriculum bounds by sampling environments from reverse ranges $\psi_i \in (\psi_t,\psi_{\mathrm{max}})$
instead of forward curriculum ranges $\psi_i \in (\psi_{\mathrm{min}},\psi_t)$.

Automatic curriculum learning and environment-generation methods such as POET \citep{wang2019poet}, PAIRED \citep{PAIRED}, REPAIRED \citep{REPAIRED}, PLR \citep{jiang2021prioritized}, and ACuTE \citep{shukla2022acute} are primarily designed for replayable levels, procedurally generated maps, or richer adversarial environment-generation settings. CLUTR \citep{CLUTR} additionally focuses on learning task representations. CURROT \citep{klink2022curriculum} operates in continuous task spaces, but represents curricula using particle-based optimal transport rather than explicit gradient-based adaptation of interpretable parameter bounds. Since the environments are defined by bounded low-dimensional continuous parameter spaces rather than procedural level generation, the curriculum is primarily compared against continuous-task curriculum methods such as ALP-GMM \citep{portelas2020teacher}, SPRL \citep{klink2020self}, and domain randomization \citep{tobin2017domain}. 

Ablations were conducted to isolate the contributions of the reparameterized critic, unidirectional curriculum updates, and MSE-based distribution-shift regularization. Specifically, the following three variants are compared:

\textbf{Reparam-A} – includes bidirectional gradient updates, allowing the curriculum to increase or decrease difficulty depending on the gradient sign. The upper bounds are updated as follows:
\begin{equation*}
\psi_{t+1}[j] =
\begin{cases}
\psi_t[j]
-
lr_{\psi[j],t}^{\mathrm{anneal}}
\left(
\eta_{\psi[j],t}
\dfrac{\partial \mathcal{L}}
{\partial \psi_t[j]}
\right),
&
\text{for all }
\dfrac{\partial \mathcal{L}}
{\partial \psi_t[j]}
< 0 \text{ or} >=0
\end{cases}
\label{eq:eq_bi}
\end{equation*}

\textbf{Reparam-M} – includes MSE regularization with unidirectional gradient updates \\
\\
\textbf{Reparam} – includes only unidirectional gradient updates without MSE regularization.

For each environment the advanced curriculum variants (Reparam-A, Reparam-R, Frontier) are based on the best-performing Reparam configuration, which includes MSE regularization when beneficial.

\subsection{Experimental Setting}
\label{sec:exp-setting}

The method is evaluated under continuously varying environment conditions during training, evaluation, and testing. In Car Racing, the environment parameter vector is $\psi=(\kappa,p)$ with feasible ranges $\kappa \in [0.31,0.71]$ and $p \in [0.05,0.13]$. In Bipedal Walker, $\psi=(f,\rho,s)$ with ranges $f \in [2.0,3.0]$, $\rho \in [0.0,0.2]$, and $s \in [0.0,0.01]$. During training, the curriculum critic adapts the running upper bounds $\psi_t$ while keeping lower bounds fixed at $\psi_{\mathrm{min}}$, gradually exposing the policy to more challenging environments. During evaluation and testing, environments are uniformly sampled from the full feasible ranges $[\psi_{\mathrm{min}},\psi_{\mathrm{max}}]$ to assess generalization.

Robustness is primarily measured by the mean episodic reward over a set $\mathbf{E}_{\mathrm{test}}$ of 500 randomly sampled environments from the full parameter ranges. All methods are trained with five random seeds to account for policy variance, and hyperparameters are tuned manually. Shared settings include PPO update frequency of 1000 timesteps, total training duration of 1M timesteps, evaluation every 10K timesteps over 10 randomly sampled $\mathbf{E}_{\mathrm{eval}}$ environments, and 10 epochs per policy update. Weighted parameter-specific learning rates relative to the critic $lr$ are applied, with an environment set size $n=200$ for stable critic gradient estimates. Experiments are conducted on an Apple M4 Pro (14-core CPU, 48 GB RAM). Additional model-specific hyperparameters and sensitivity analyses are provided in the Appendix.

\subsection{Results}
\label{sec:results} 

\begin{figure*}
    \centering
    \begin{subfigure}{0.45\linewidth}
        \centering
        \includegraphics[width=\linewidth]{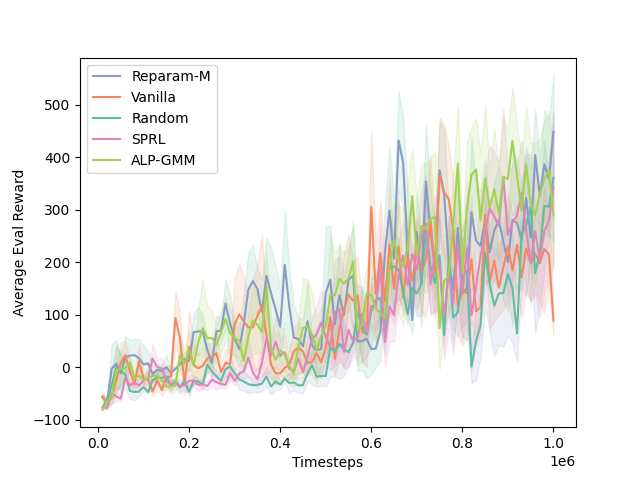}
        \caption{Average evaluation rewards- Car Racing}
        \label{fig1a}
    \end{subfigure}
    \hfill
    \begin{subfigure}{0.45\linewidth}
        \centering
        \includegraphics[width=\linewidth]{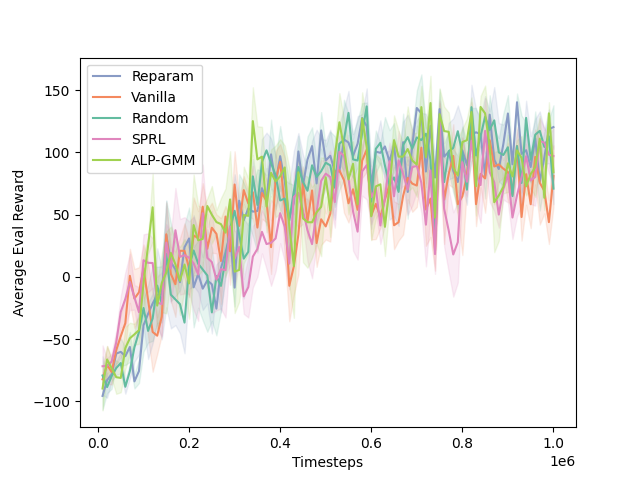}
        \caption{Average evaluation rewards- Bipedal Walker}
        \label{fig1b}
    \end{subfigure}

    \caption{Mean Evaluation performance measured every 10000 training timesteps; shaded regions indicate ±1 standard error of the mean.}
    \label{fig4}
\end{figure*}

\begin{figure}[htbp]
    \centering

    \begin{subfigure}{0.39\textwidth}
        \centering
        \includegraphics[width=\linewidth]{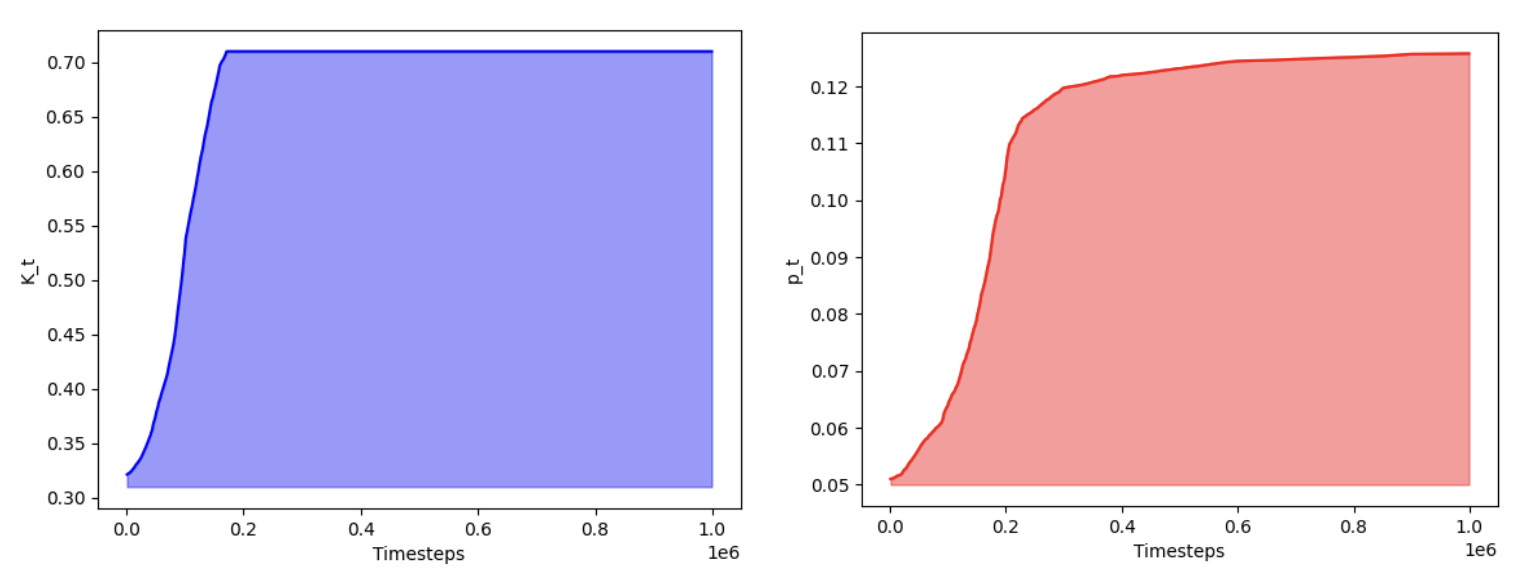}
        \caption{$\kappa_i \in (0.31, \kappa_t)$, $p_i \in (0.05, p_t)$}
    \end{subfigure}
    \hfill
    \begin{subfigure}{0.595\textwidth}
        \centering
        \includegraphics[width=\linewidth]{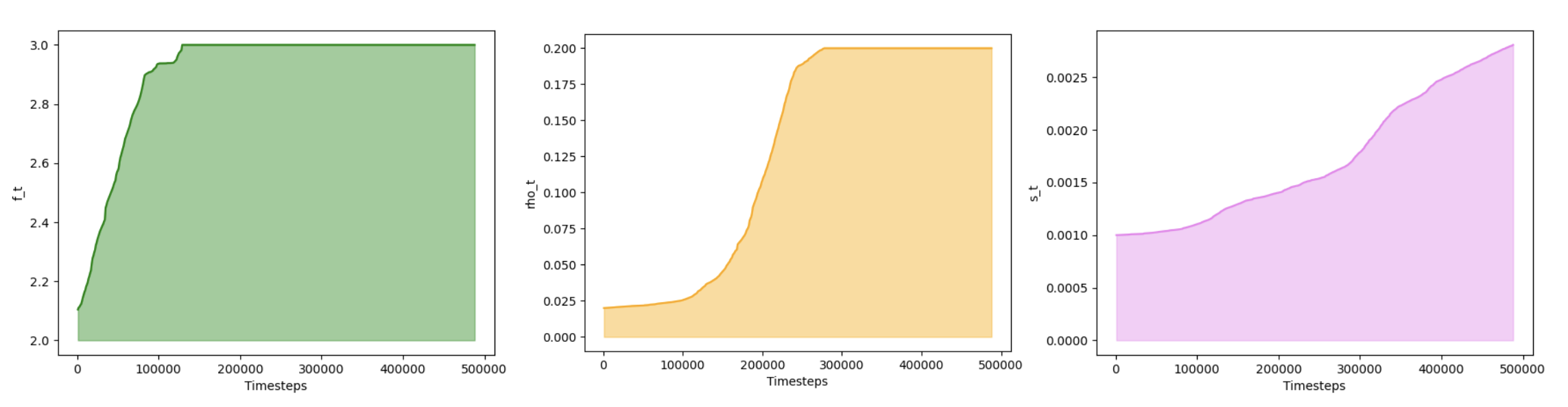}
        \caption{$f_i \in (0.2, f_t)$, $\rho_i \in (0.0, \rho_t)$, $s_i \in (0.0, s_t)$}
    \end{subfigure}

    \caption{Curriculum Growth Curves in a) Car Racing and b) Bipedal Walker}
    \label{fig5}
\end{figure}

Results are presented on the $\mathbf{E}_{\mathrm{test}}$ environments (Section \ref{sec:exp-setting}). Training curves (Fig.~\ref{fig4}) report the mean episodic reward evaluated every 10K timesteps over 10 sampled $\mathbf{E}_{\mathrm{eval}}$ environments per seed, averaged across five training seeds. Final testing performance (Tables \ref{table1} and \ref{table2}) is obtained by selecting the best checkpoint from each seed based on evaluation performance during training, and re-evaluating over 500 sampled $\mathbf{E}_{\mathrm{test}}$ environments. The mean $\pm$ standard deviation are reported together with the additional metrics from Section \ref{sec:metrics}. Curriculum parameter evolution over training is shown in Fig.~\ref{fig5}.

Results in Tables 1 and 2 indicate that \textbf{Reparam-M} achieves the best performance in Car Racing, while \textbf{Reparam} performs best in Bipedal Walker. This shows that MSE-based distribution-shift regularization improves stability and performance for the multimodal Nature CNN + MLP policy in Car Racing, but provides limited benefit for the lower-dimensional MLP-only Bipedal Walker policy. These observations suggest that the auxiliary regularization is particularly useful in image-based settings, where distribution shifts have a greater impact on learned latent representations. In contrast, bidirectional gradient updates in \textbf{Reparam-A} do not improve performance, as allowing the curriculum to decrease difficulty when value is positive slows down overall learning progress.

Since environments are sampled from the running bounds $(\psi_{\mathrm{min}}, \psi_t)$, the reparameterization method gradually expands environment difficulty while continuing to expose the policy to easier environments, helping reduce catastrophic forgetting. Figure~\ref{fig4} further shows that it achieves higher evaluation rewards than the baselines toward later stages of training.

\begin{table*}[h!]   
\caption{Average Testing Performance across 500 evaluation environments (5 training seeds)- Car Racing. } 
\begin{center}
\begin{tabular*}{0.965\linewidth}{|c|c|c|c|c|c|c|}
\hline 
\textbf{Training} & \textbf{Training} & \textbf{Average} & \textbf{Number of} & \textbf{Collision} & \textbf{Tiles} & \textbf{Time on}  \\ 
\textbf{Scheme} & \textbf{Time (hrs)} & \textbf{Reward} & \textbf{Collisions} & \textbf{Obs Ratio} & \textbf{Visited} & \textbf{Grass} \\
\hline
 Vanilla & 3.825 &
$540 \pm 178$ & $1.8484$ & $0.4234$ & $213$ & $0.3611$ \\ 
 \hline
 Random & 3.798 & $557 \pm 174$ & $1.1752$ & $0.1429$ & $208$ & $0.3737$ \\
 \hline
 Manual & 3.729 & $617 \pm 166$ & $\mathbf{0.3892}$ & $\mathbf{0.0470}$ & $214$ & $0.20737$ \\
 \hline
 SPRL & \textbf{3.669} & $528 \pm 133$ & $0.846$ & $0.10189$ & $195$ & $0.2810$ \\
 \hline
 ALP-GMM & 3.716 & $508 \pm 148$ & $1.078$ & $0.1295$ & $192$ & $0.7895$ \\
 \hline
 Frontier & 4.653 & $500 \pm 160$ & $0.528$ & $0.0634$ & $182$ & $0.4826$ \\
 \hline
 Reparam-R & 4.596 & $542 \pm 147$ & $0.6516$ & $0.0779$ & $197$ & $0.5633$ \\
 \hline
 \textbf{Reparam-M} & $4.714$ & $\mathbf{650 \pm 134}$ & $0.601$ & $0.0729$ & $\mathbf{227}$ & $0.2449$ \\
 \hline
 \multicolumn{7}{|c|}{Ablations} \\ 
 \hline
 Reparam & 4.5912 & $448 \pm 122$ & $0.5784$ & $0.0691$ & $167$ & $1.8453$ \\
 \hline
 Reparam-A & 4.548 & $623 \pm 153$ & $0.5484$ & $0.0648$ & $218$ & $\mathbf{0.2059}$ \\
 \hline
\end{tabular*}
\end{center}
\label{table1}
\end{table*}

\begin{table*}[h!]   
\caption{Average Testing Performance across 500 evaluation environments (5 training seeds)- Bipedal Walker. } 
\begin{center}
\begin{tabular*}{0.95\linewidth}{|c|c|c|c|c|c|c|}
\hline 
\textbf{Training} & \textbf{Training} & \textbf{Average} & \textbf{Pit Fall} & \textbf{Slope Fail-} & \textbf{Slip Fail-} & \textbf{Steps}  \\ 
\textbf{Scheme} & \textbf{Time (min)} & \textbf{Reward} & \textbf{Rate} & \textbf{ure Rate} & \textbf{ure Rate} & \textbf{Covered} \\
\hline
Vanilla & \textbf{9.638} & $90 \pm 144$  & $\textbf{0.0723}$ & $0.362$ & $0.351$ & $110$ \\
\hline
Random & 10.552 & $111 \pm 132$  & $0.0997$ & $0.3944$ & $0.3852$ & $125$ \\
\hline
Manual & 10.219 & $117 \pm 142$  & $0.0855$ & $\mathbf{0.325}$ & $0.311$ & $125$ \\
\hline
SPRL & 10.526 & $88 \pm 142$  & $0.1068$ & $0.380$ & $0.386$ & $110$\\
\hline
ALP-GMM & 10.394 & $116 \pm 139$  & $0.0916$ & $0.327$ & $\mathbf{0.310}$ & $127$ \\
\hline
Frontier & 10.772 & $93 \pm 139$  & $0.1111$ & $0.368$ & $0.399$ & $113$ \\
\hline
Reparam-R & 11.358 & $109 \pm 128$  & $0.1022$ & $0.418$ & $0.422$ & $124$\\
\hline
\textbf{Reparam} & 10.428 & $\mathbf{127 \pm 136}$ & $0.1093$ & $0.379$ & $0.374$ & $\mathbf{128}$ \\
\hline
\multicolumn{7}{|c|}{Ablations} \\ 
 \hline
 Reparam-M & 10.514 & $112 \pm 131$ & $0.1076$ & $0.367$ & $0.374$ & $122$ \\
 \hline
 Reparam-A & 10.529 & $110 \pm 137$ & $0.141$ & $0.336$ & $0.327$ & $123$ \\
\hline
\end{tabular*}
\end{center}
\label{table2}
\end{table*}

\section{Conclusion}
\label{sec:conclusion}

In this paper, a sample-efficient curriculum generation method is presented based on reparameterization and auxiliary distribution-shift regularization to train robust RL navigation policies in continuous environment parameter spaces. It demonstrated competitive performance against adaptive curriculum baselines while balancing metrics across varying environment conditions. Future work includes scaling to higher-dimensional parameter spaces representative of real-world variability.



\bibliography{example}  

\clearpage

\appendix

\section*{\centering \huge{Appendix}}

\section{Policy-Critic Architecture}

\begin{figure*}[h]
    \centering
    \includegraphics[scale=0.27]{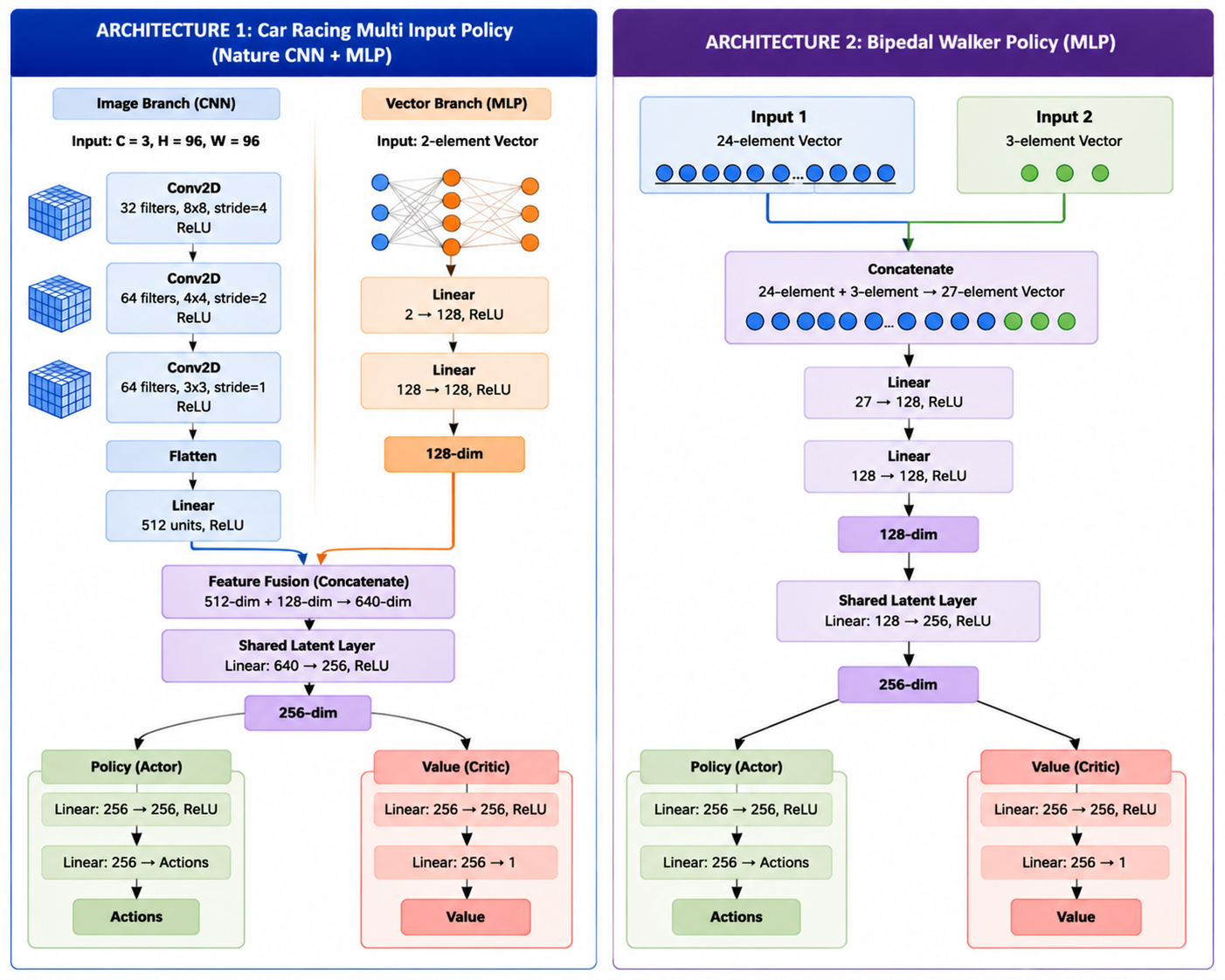}
    \caption{Architecture of the Policy and Critic Networks in Car Racing and Bipedal Walker Environments.}
    \label{fig6}
\end{figure*}

Figure \ref{fig6} illustrates the architectures of the policy and critic networks for the Car Racing and Bipedal Walker environments. Before policy and curriculum optimization, raw environment observations are processed into appropriate feature representations. For Car Racing, a multimodal Nature CNN + MLP network is employed: $96\times96\times3$ RGB track images $x$ are processed through a CNN, while the 2-element environment parameters $\psi=(\kappa,p)$ are encoded via an MLP; the resulting features are then fused for policy learning. For Bipedal Walker, 24-element proprioceptive state vectors $x$ are concatenated with 3-element environment vectors $\psi=(f,\rho,s)$ and processed through an MLP. The policy is trained using Proximal Policy Optimization (PPO) with clipped policy and value objectives. Additionally, the value network is leveraged to compute gradients with respect to the running curriculum bounds $\psi_t$, which are used to adaptively adjust environment difficulty.

\section{Curriculum Update Rule Derivation}
\label{sec:curriculum_derivation}

The critic adapts the curriculum by estimating the expected value of the policy across environments sampled within the current curriculum bounds. Let $\psi_\mathrm{min}$ and $\psi_t$ denote the lower and current upper bounds of the curriculum, respectively. Environments are initially sampled uniformly as:
\begin{equation*}
\psi_i \sim U[\psi_\mathrm{min}, \psi_t].
\end{equation*}

To enable gradient-based curriculum updates, the sampled environments are reparameterized:
\begin{equation*}
\psi_i = \psi_\mathrm{min} + z_i \odot (\psi_t - \psi_\mathrm{min}), \qquad z_i \sim U[0,1],
\end{equation*}
where $\odot$ denotes element-wise multiplication. For the $j^{\mathrm{th}}$ parameter:
\begin{equation*}
\psi_i[j] = \psi_\mathrm{min}[j] + z_i[j] (\psi_t[j]-\psi_\mathrm{min}[j]), \quad
\frac{\partial \psi_i[j]}{\partial \psi_t[j]} = z_i[j].
\end{equation*}

The critic objective $\mathcal{L}$ is computed over $n$ sampled environments, optionally including a distribution-shift regularization term to constrain abrupt changes between consecutive curriculum distributions:
\begin{equation}
\mathcal{L} = \frac{1}{n} \sum_i \Big( V_\phi(x_i, \psi_i) - \beta \delta(\psi_i, \psi_i^\mathrm{prev}) \Big),
\label{eq:critic_objective}
\end{equation}
where $V_\phi$ is the PPO value network, $\delta(\psi_i, \psi_i^\mathrm{prev}) = \|\psi_i - \psi_i^\mathrm{prev}\|^2$, and $\beta \ge 0$ controls the strength of the regularization.

For the $j^{\mathrm{th}}$ environment parameter, the gradient of the critic objective with respect to the running curriculum bound $\psi_t[j]$ is:
\begin{align*}
\frac{\partial \mathcal{L}}{\partial \psi_{\mathrm{t}}[j]} &= \frac{1}{n} \frac{\partial}{\partial \psi_{\mathrm{t}}[j]} (\sum_\mathrm{i}^{n} V_{\phi}(x_\mathrm{i},\psi_\mathrm{i})- \beta \delta(\psi_\mathrm{i}, \psi^{\mathrm{prev}}_\mathrm{i})) \\ &= \frac{1}{n} \sum_\mathrm{i}^{n} \frac{\partial V_{\phi}(x_\mathrm{i},\psi_\mathrm{i})}{\partial \psi_{\mathrm{t}}[j]} - \beta\frac{\partial}{\partial \psi_{\mathrm{t}}[j]}||\psi_\mathrm{i}- \psi^{\mathrm{prev}}_\mathrm{i}||^2 \\
&= \frac{1}{n} \sum_i^{n} \Bigg( 
\frac{\partial V_\phi(x_i, \psi_i)}{\partial \psi_i[j]} \frac{\partial \psi_i[j]}{\partial \psi_t[j]} 
- 2 \beta (\psi_i[j] - \psi_i^\mathrm{prev}[j]) \frac{\partial \psi_i[j]}{\partial \psi_t[j]} 
\Bigg) \\
&= \frac{1}{n} \sum_i^{n} \Big( 
\frac{\partial V_\phi(x_i, \psi_i)}{\partial \psi_i[j]} - 2 \beta (\psi_i[j] - \psi_i^\mathrm{prev}[j])
\Big) z_i[j].
\end{align*}
Negative values of $\frac{\partial \mathcal{L}}{\partial \psi_t[j]}$ indicate that increasing $\psi_t[j]$ would reduce the estimated policy value, guiding the curriculum to expose the policy to more difficult environments. The curriculum is updated as:
\begin{equation}
\psi_{t+1}[j] =
\begin{cases}
\psi_t[j] - lr_{\psi[j],t}^{\mathrm{anneal}} \, \eta_{\psi[j],t} \, \frac{\partial \mathcal{L}}{\partial \psi_t[j]}, & \text{if } \frac{\partial \mathcal{L}}{\partial \psi_t[j]} < 0 \\
\psi_t[j], & \text{otherwise}
\end{cases}
\label{eq:curriculum_update}
\end{equation}
where the annealed learning rate $lr_{\psi[j],t}^{\mathrm{anneal}}$ and normalization constant $\eta_{\psi[j],t}$ are defined as:
\begin{equation}
lr_{\psi[j],t}^{\mathrm{anneal}} = \frac{lr_{\psi[j]}}{\alpha_t}
\label{eq5}
\end{equation}

\begin{equation}
\eta_{\psi[j],t} = \frac{\psi_t[j]-\psi_\mathrm{min}[j]}{\psi_\mathrm{max}[j]-\psi_\mathrm{min}[j]}.
\label{eq6}
\end{equation}

This ensures that curriculum bounds increase only when doing so introduces more challenging environments, while leaving other parameters unchanged.

Additionally for the Reparam-A ablation, the curriculum effects when all gradients (positive or negative) are demonstrated as follows:

\begin{equation*}
\psi_{t+1}[j] =
\begin{cases}
\psi_t[j]
-
lr_{\psi[j],t}^{\mathrm{anneal}}
\left(
\eta_{\psi[j],t}
\dfrac{\partial \mathcal{L}}
{\partial \psi_t[j]}
\right),
&
\text{for all }
\dfrac{\partial \mathcal{L}}
{\partial \psi_t[j]}
< 0 \text{ or} >=0
\end{cases}
\end{equation*}

\section{Training Hyperparameters}

The experiments were conducted on five random seeds for all baselines and ablations to account for inherent variance in reinforcement learning across training seeds and hyperparameter tuning was performed using a manual search. Details of the model-specific hyperparameters can be referred in Tables \ref{table3} and \ref{table4} .

\begin{table}[h!]   
\caption{Model-Specific Hyperparameters- Car Racing}
\begin{center}
\begin{tabular}{ |c|c|c|c|c|c|c|c|c| } 
\hline
Model & Policy $lr$ & Init $\kappa,p$ bounds & $n$ & Critic $lr$ & $\kappa$:$p$ $lr$ ratio  & $\beta$ \\ 
 \hline
 Vanilla & 6e-4 & - & - & - & - & - \\ 
 \hline
  Random & 5e-4 & - & 200 & - & - & - \\ 
  \hline
  Manual & 5e-4 & -  & 200 & - & - & - \\ 
 \hline
 SPRL & 5e-4 & - &  200 & - & - & - \\ 
 \hline
 ALP-GMM & 5e-4 & - &  200 & - & - & - \\ 
 \hline
  Reparam & 5e-4 & (0.31,0.32), (0.05,0.051) & 200 & 2e-4 & 1:8 & 0.0 \\ 
  \hline
  Reparam-M & 5e-4 & (0.31,0.32), (0.05,0.051) & 200 & 2e-4 & 1:8 & 1.0 \\ 
  \hline
  Reparam-A & 5e-4 & (0.31,0.32), (0.05,0.051) & 200 & 2e-4 & 1:8 & 1.0 \\ 
  \hline
  Reparam-R & 5e-4 & (0.70,0.71), (0.129,0.13) & 200 & 2e-4 & 1:8 & 1.0 \\ 
  \hline
 Frontier & 5e-4 & (0.31,0.32), (0.05,0.051) & 200 & 2e-4 & 1:8 & 1.0  \\ 
 \hline
\end{tabular}
\label{table3}
\end{center}
\end{table}

\begin{table}[h!]   
\caption{Model-Specific Hyperparameters- Bipedal Walker}
\begin{center}
\begin{tabular}{ |c|c|c|c|c|c|c|c|c| } 
\hline
Model & Policy $lr$ & Init $f,\rho,s$ bounds & $n$ & Critic $lr$ & $f$:$\rho$:$s$ $lr$ ratio  & $\beta$ \\ 
 \hline
 Vanilla & 3e-4 & - & - & - & - & - \\ 
 \hline
  Random & 3e-4 & - &  200 & - & - & - \\ 
  \hline
  Manual & 3e-4 & - &  200 & - & - & - \\ 
 \hline
 SPRL & 3e-4 & - & 200 & - & - & -  \\ 
 \hline
 ALP-GMM & 3e-4 & - &  200 & - & -  & - \\ 
 \hline
  Reparam & 3e-4 & (2.0, 2.1), (0.0,0.02), (0.0, 0.001) &  200 & 2e-4 & 2:4:1 & 0.0 \\ 
  \hline
  Reparam-M & 3e-4 & (2.0,2.1), (0.0,0.02), (0.0, 0.001) &  200 & 2e-4 & 2:4:1 & 1.0 \\ 
  \hline
  Reparam-A & 3e-4 & (2.0,2.1), (0.0,0.02), (0.0, 0.001) & 200 & 2e-4 & 2:4:1 & 0.0 \\ 
  \hline
  Reparam-R & 3e-4 & (2.9,3.0), (0.18,0.20), (0.009, 0.01) &  200 & 2e-4 & 2:4:1 & 0.0 \\ 
  \hline
 Frontier & 3e-4 & (2.0,2.1), (0.0,0.02), (0.0, 0.001) &  200 & 2e-4 & 2:4:1 & 0.0  \\ 
 \hline
\end{tabular}
\label{table4}
\end{center}
\end{table}

\section{Results- Car Racing}

Below, the training evaluation curves and curriculum growth curves are presented for all experiments reported in the main paper for the Car Racing environment, with all curves averaged over five random seeds (0,1,2,3,4). Additionally, the interquartile mean (IQM) and per-seed average rewards are reported for each method.

\begin{figure*}[htbp!]
    \centering
    \begin{subfigure}{0.425\linewidth}
        \centering
        \includegraphics[width=\linewidth]{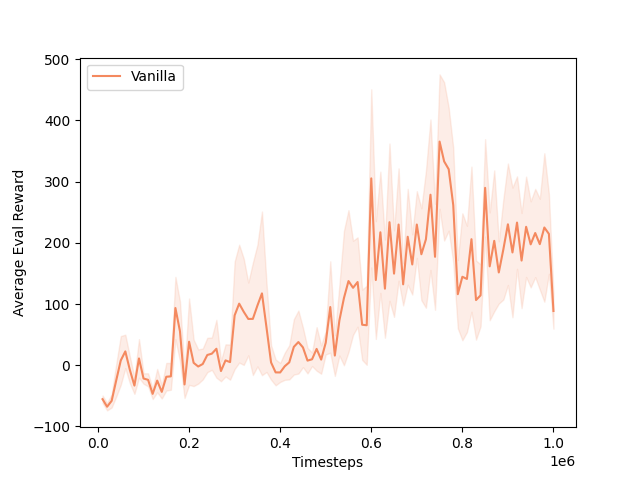}
        \caption*{Vanilla Policy}
        \label{fig1a}
    \end{subfigure}
    \hfill
    \begin{subfigure}{0.45\linewidth}
        \centering
        \includegraphics[width=\linewidth]{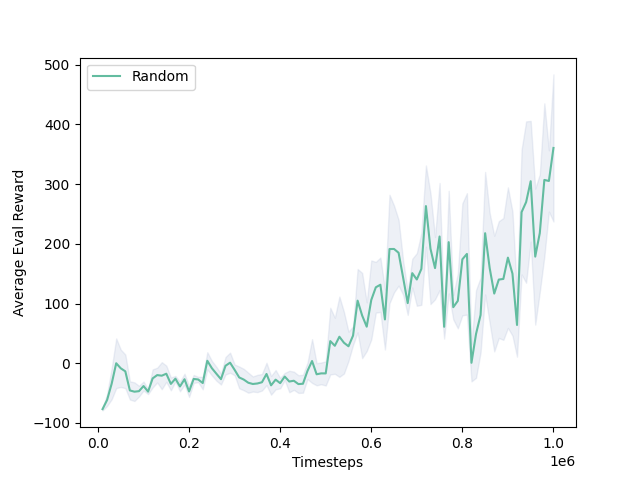}
        \caption*{Random Sampling}
        \label{fig1b}
    \end{subfigure}
\end{figure*}

\begin{figure*}
    \begin{subfigure}{0.45\linewidth}
        \centering
        \includegraphics[width=\linewidth]{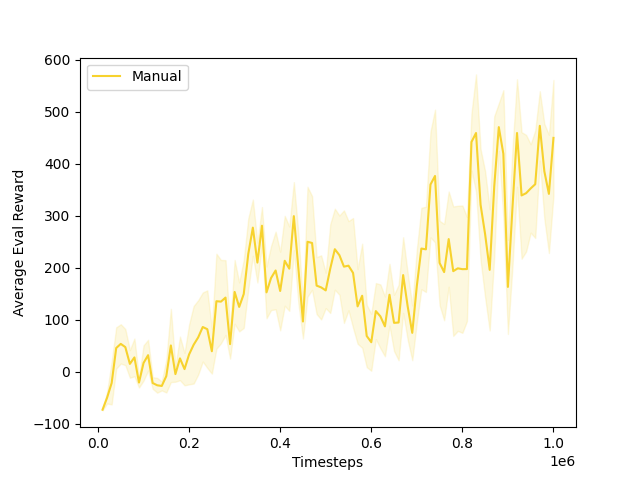}
        \caption*{Manual Curriculum}
        \label{fig2a}
    \end{subfigure}
    \hfill
    \begin{subfigure}{0.45\linewidth}
        \centering
        \includegraphics[width=\linewidth]{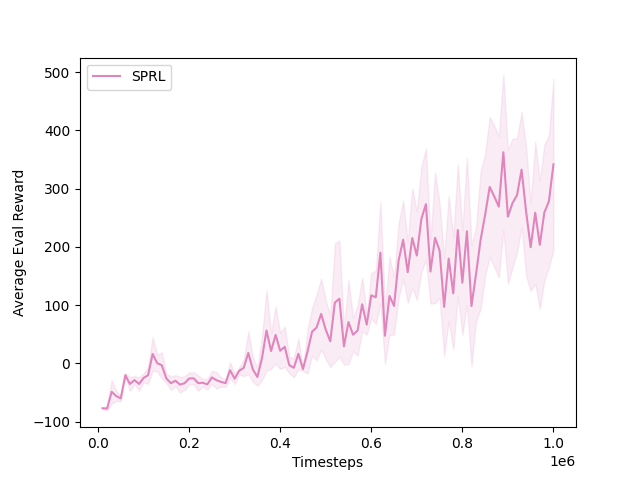}
        \caption*{Self-Paced RL}
        \label{fig2b}
    \end{subfigure}

    \begin{subfigure}{0.45\linewidth}
        \centering
        \includegraphics[width=\linewidth]{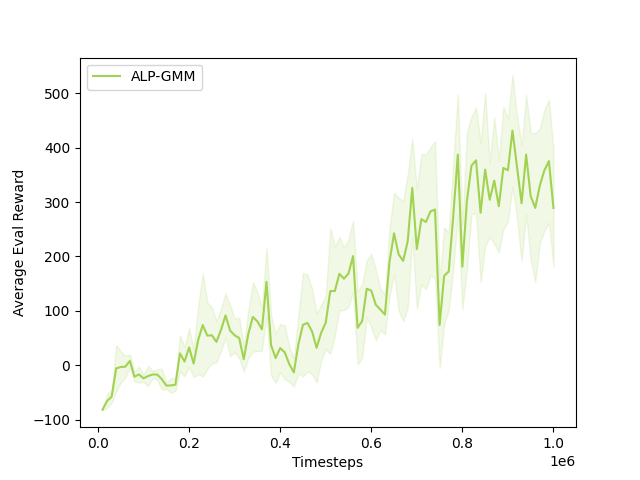}
        \caption*{ALP-GMM}
        \label{fig3a}
    \end{subfigure}
    \hfill
    \begin{subfigure}{0.45\linewidth}
        \centering
        \includegraphics[width=\linewidth]{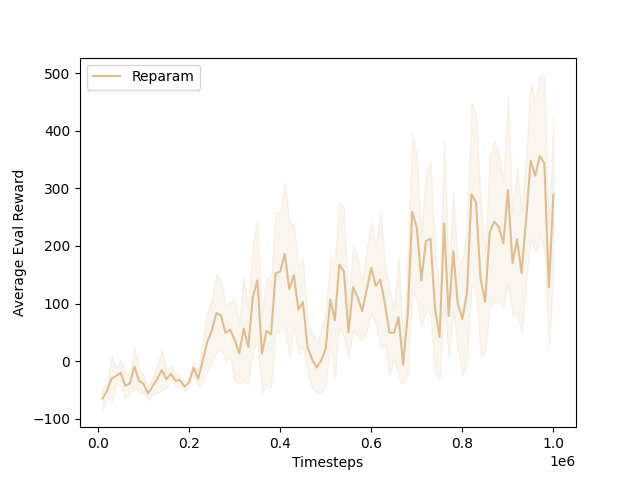}
        \caption*{Reparam}
        \label{fig3b}
    \end{subfigure}

    \begin{subfigure}{0.45\linewidth}
        \centering
        \includegraphics[width=\linewidth]{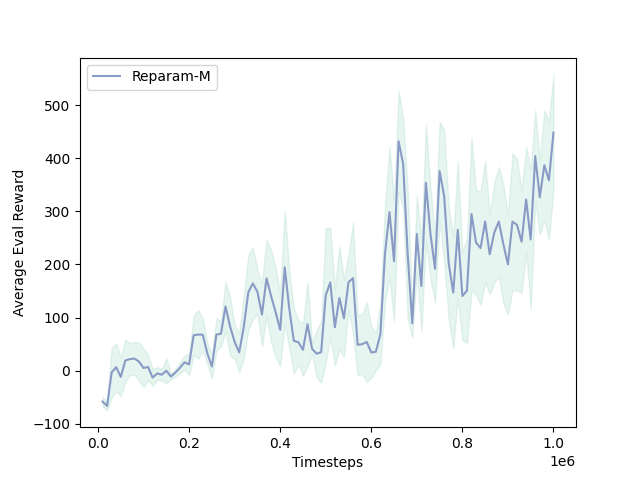}
        \caption*{Reparam-M}
        \label{fig4a}
    \end{subfigure}
    \hfill
    \begin{subfigure}{0.45\linewidth}
        \centering
        \includegraphics[width=\linewidth]{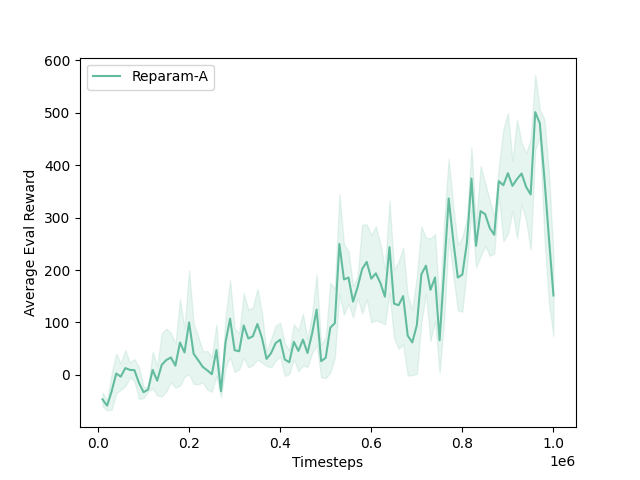}
        \caption*{Reparam-A}
        \label{fig4b}
    \end{subfigure}

    \begin{subfigure}{0.45\linewidth}
        \centering
        \includegraphics[width=\linewidth]{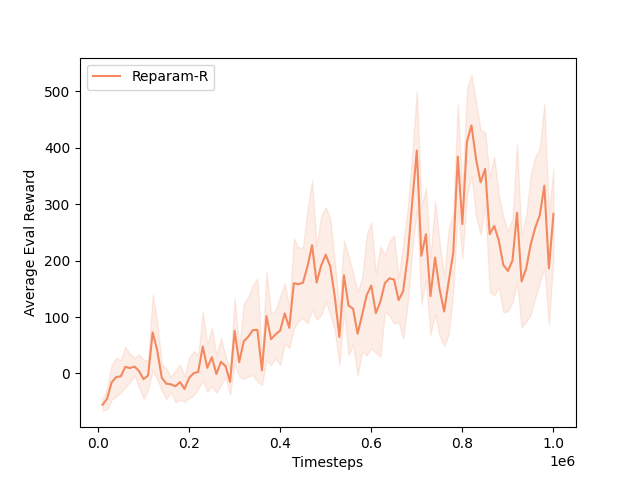}
        \caption*{Reparam-R}
        \label{fig4a}
    \end{subfigure}
    \hfill
    \begin{subfigure}{0.45\linewidth}
        \centering
        \includegraphics[width=\linewidth]{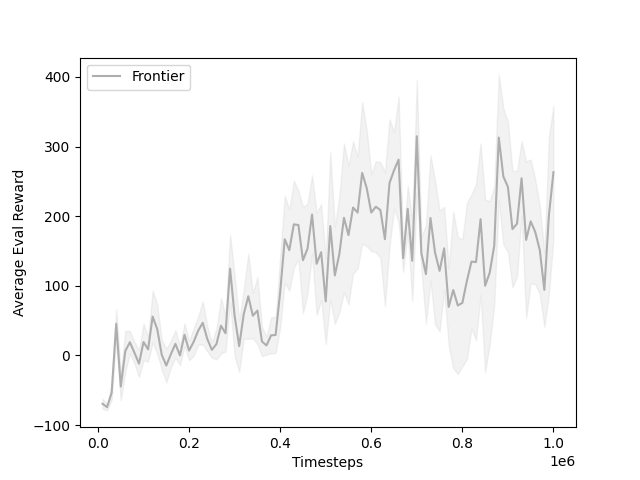}
        \caption*{Frontier}
        \label{fig4b}
    \end{subfigure}
    \caption*{Figure 2: Mean evaluation performance measured every 10000 training timesteps. Curves show the average evaluation reward across 5 random seeds; shaded regions indicate ±1 standard error of the mean. Each evaluation consists of rollouts over 10 environments.- Car Racing}
\end{figure*}

\begin{figure}[htbp]
    \centering
    \begin{subfigure}{0.45\textwidth}
        \centering
        \includegraphics[width=\linewidth,height=4cm]{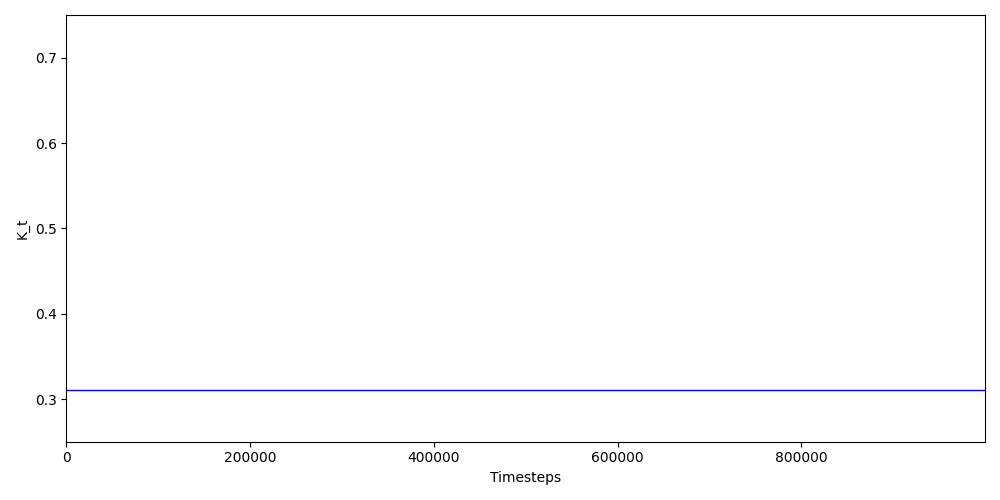}
        \caption*{Vanilla Policy- No Curriculum $\kappa_t=0.31$}
    \end{subfigure}
    \hfill
    \begin{subfigure}{0.43\textwidth}
        \centering
        \includegraphics[width=\linewidth,height=4cm]{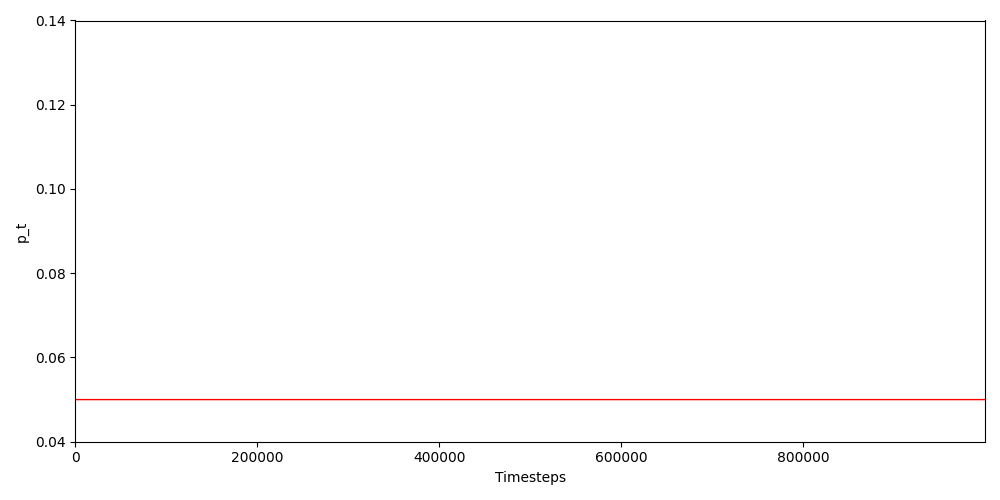}
        \caption*{Vanilla Policy- No Curriculum $p_t=0.05$}
    \end{subfigure}
    
    \begin{subfigure}{0.45\textwidth}
        \centering
        \includegraphics[width=\linewidth,height=5cm]{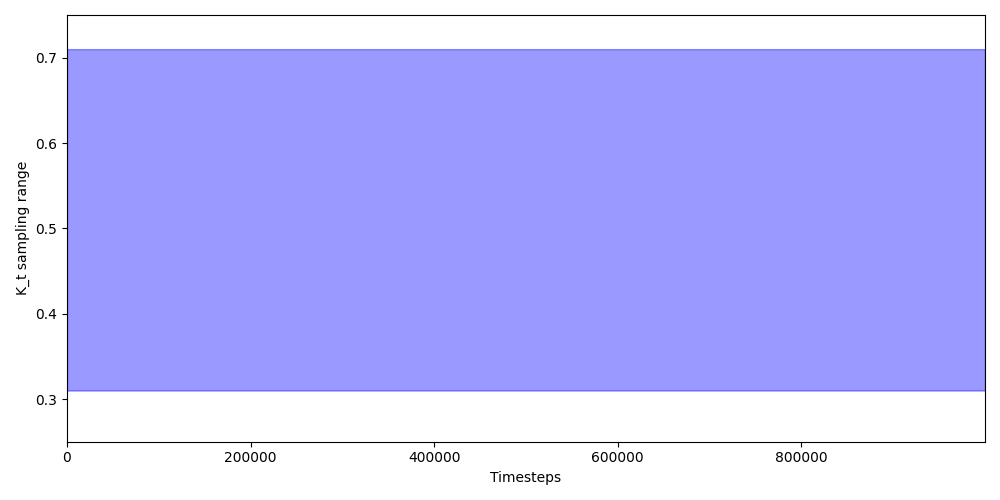}
        \caption*{Random Sampling $\kappa_t \in (0.31, 0.71)$}
    \end{subfigure}
    \hfill
    \begin{subfigure}{0.45\textwidth}
        \centering
        \includegraphics[width=\linewidth,height=5cm]{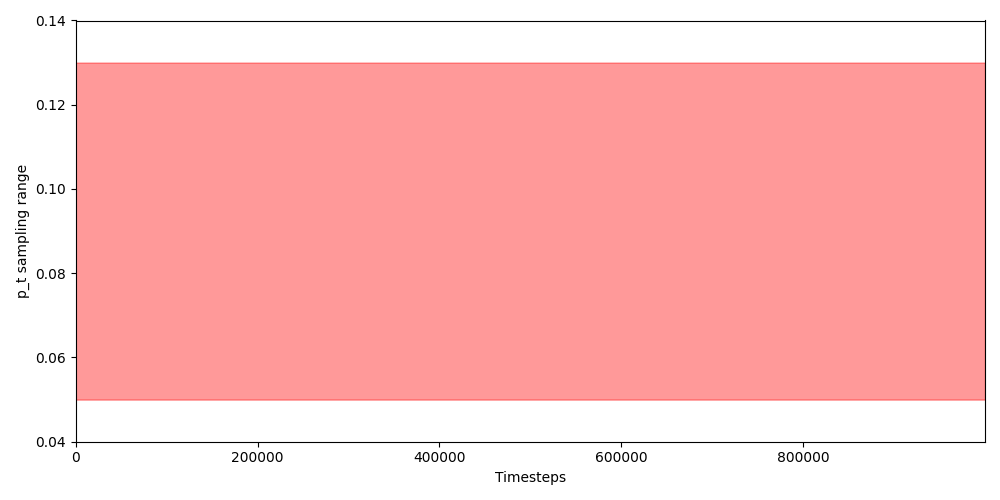}
        \caption*{Random Sampling $p_t \in (0.05, 0.13)$}
    \end{subfigure}
    
    \begin{subfigure}{0.48\textwidth}
        \centering
        \includegraphics[width=\linewidth]{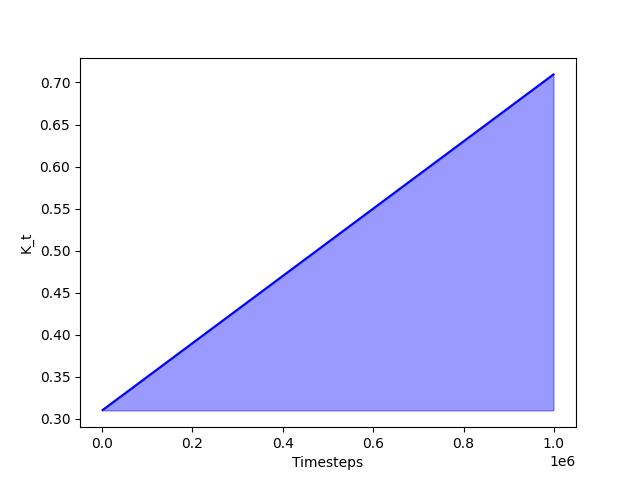}
        \caption*{Manual curriculum $\kappa_i \in (0.31, \kappa_t)$}
    \end{subfigure}
    \hfill
    \begin{subfigure}{0.48\textwidth}
        \centering
        \includegraphics[width=\linewidth]{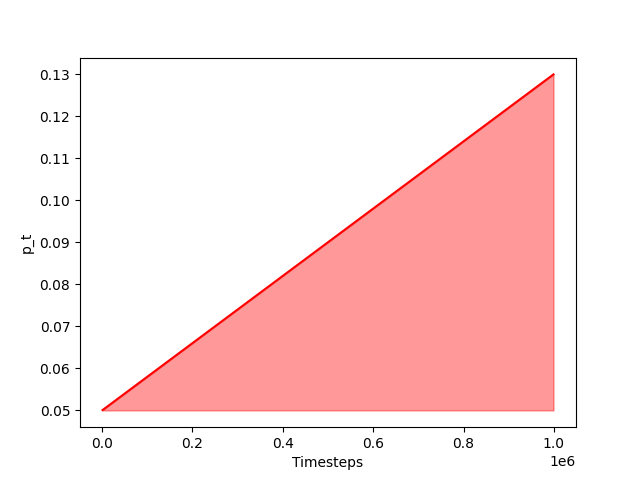}
        \caption*{Manual curriculum $p_i \in (0.05, p_t)$}
    \end{subfigure}

    \begin{subfigure}{0.48\textwidth}
        \centering
        \includegraphics[width=\linewidth]{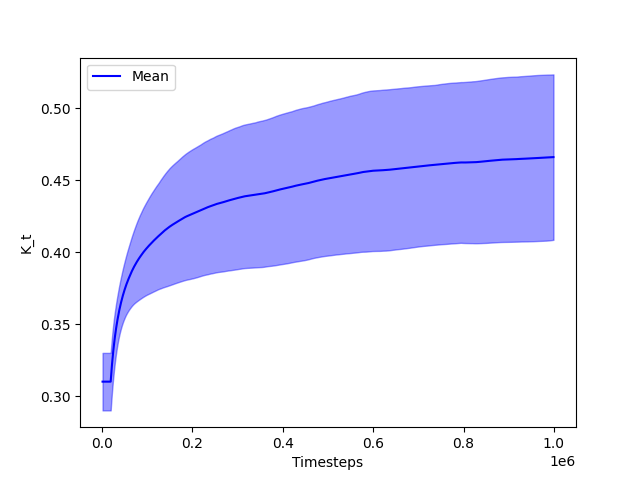}
        \caption*{Self-paced RL $\kappa_i \sim \mathcal{N}(\mu_t^\kappa,\Sigma_t^\kappa)$}
    \end{subfigure}
    \hfill
    \begin{subfigure}{0.48\textwidth}
        \centering
        \includegraphics[width=\linewidth]{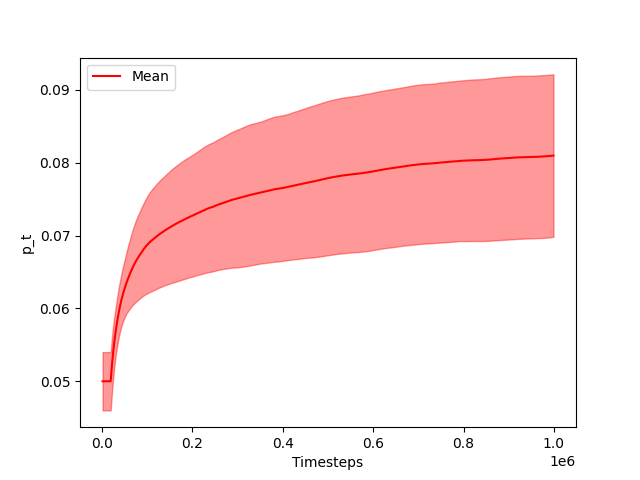}
        \caption*{Self-paced RL $p_i \sim \mathcal{N}(\mu_t^p,\Sigma_t^p)$}
    \end{subfigure}

\end{figure}

\begin{figure}

     \begin{subfigure}{0.48\textwidth}
        \centering
        \includegraphics[width=\linewidth]{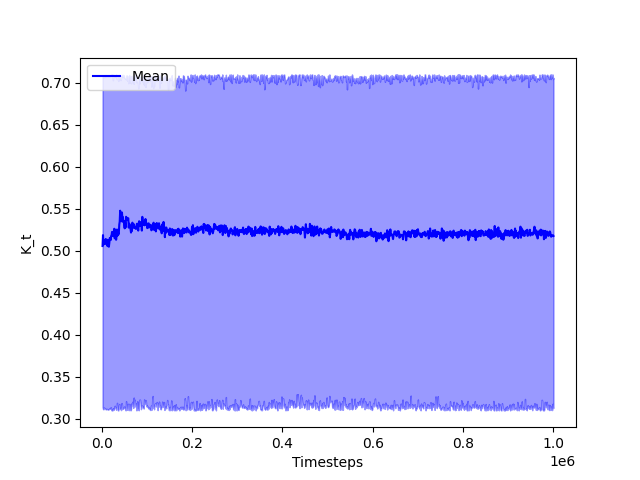}
        \caption*{ALP-GMM $\kappa_i \sim \sum_{k=1}^{K} w_k \mathcal{N}(\mu_k^\kappa,\Sigma_k^\kappa)$}
    \end{subfigure}
    \hfill
    \begin{subfigure}{0.48\textwidth}
        \centering
        \includegraphics[width=\linewidth]{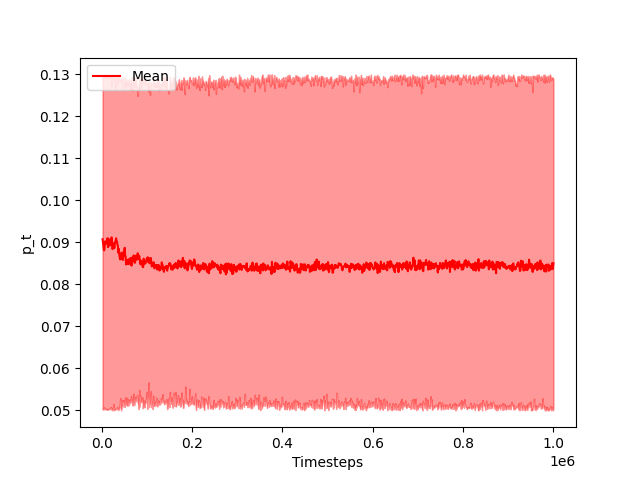}
        \caption*{ALP-GMM $p_i \sim \sum_{k=1}^{K} w_k \mathcal{N}(\mu_k^p,\Sigma_k^p)$}
    \end{subfigure}

    \begin{subfigure}{0.48\textwidth}
        \centering
        \includegraphics[width=\linewidth]{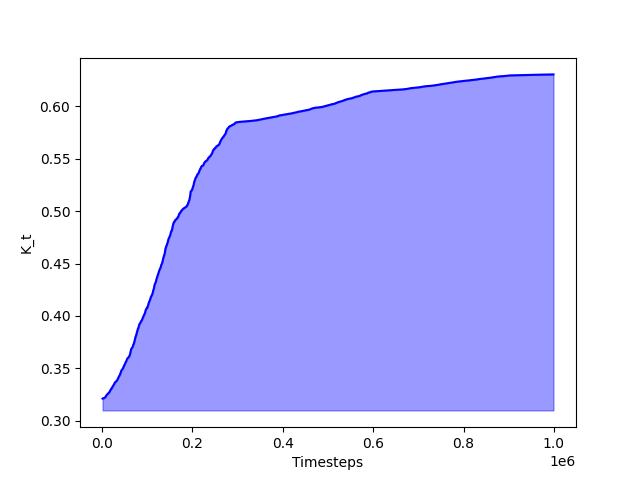}
        \caption*{Reparam curriculum $\kappa_i \in (0.31, \kappa_t)$}
    \end{subfigure}
    \hfill
    \begin{subfigure}{0.48\textwidth}
        \centering
        \includegraphics[width=\linewidth]{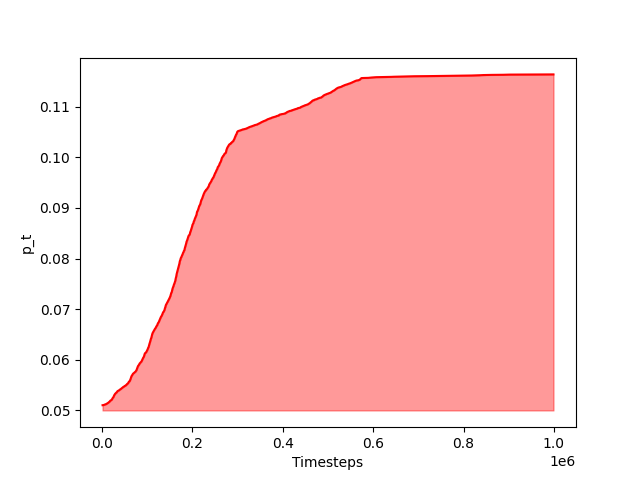}
        \caption*{Reparam curriculum $p_i \in (0.05, p_t)$}
    \end{subfigure}
    
    \begin{subfigure}{0.48\textwidth}
        \centering
        \includegraphics[width=\linewidth]{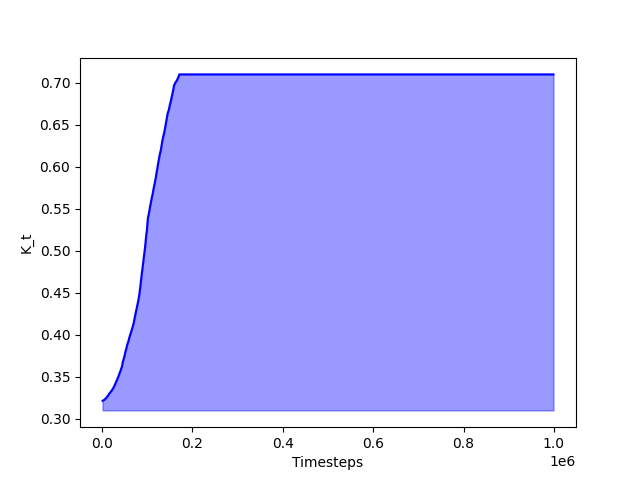}
        \caption*{Reparam-M curriculum $\kappa_i \in (0.31, \kappa_t)$}
    \end{subfigure}
    \hfill
    \begin{subfigure}{0.48\textwidth}
        \centering
        \includegraphics[width=\linewidth]{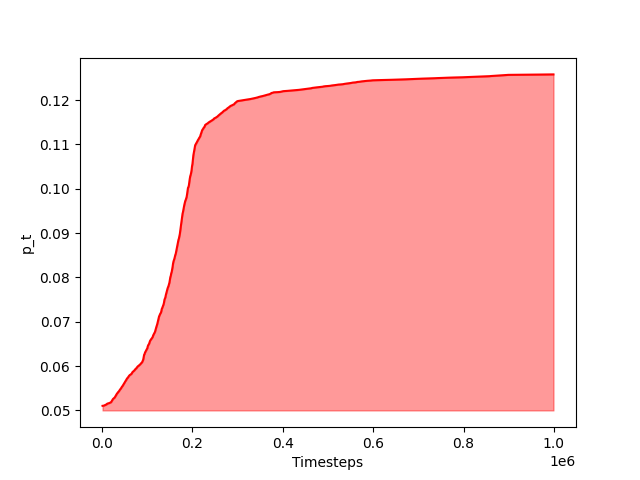}
        \caption*{Reparam-M curriculum $p_i \in (0.05, p_t)$}
    \end{subfigure}

    \begin{subfigure}{0.48\textwidth}
        \centering
        \includegraphics[width=\linewidth]{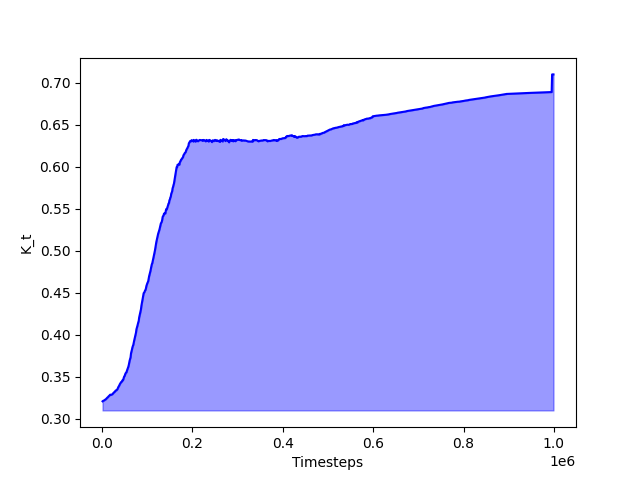}
        \caption*{Reparam-A curriculum $\kappa_i \in (0.31, \kappa_t)$}
    \end{subfigure}
    \hfill
    \begin{subfigure}{0.48\textwidth}
        \centering
        \includegraphics[width=\linewidth]{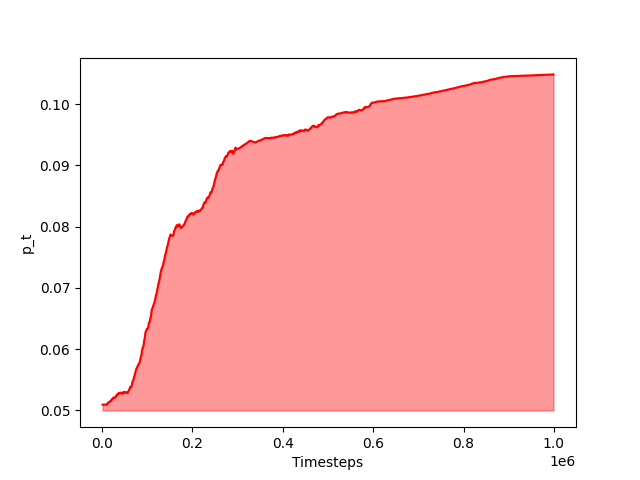}
        \caption*{Reparam-A curriculum $p_i \in (0.05, p_t)$}
    \end{subfigure}
\end{figure}

\begin{figure}
    \begin{subfigure}{0.48\textwidth}
        \centering
        \includegraphics[width=\linewidth]{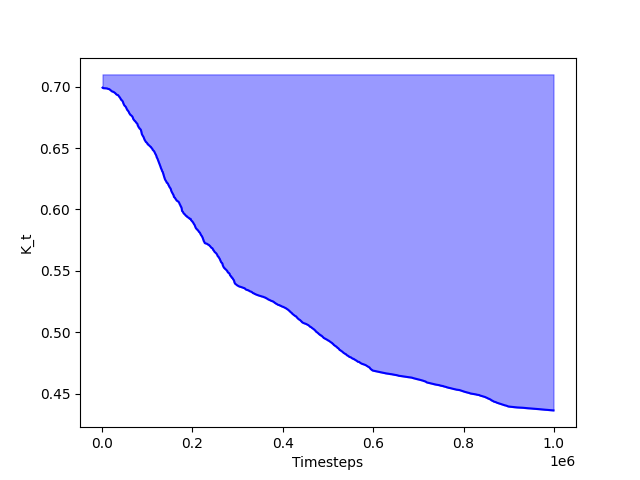}
        \caption*{Reparam-R curriculum $\kappa_i \in (\kappa_t, 0.71)$}
    \end{subfigure}
    \hfill
    \begin{subfigure}{0.48\textwidth}
        \centering
        \includegraphics[width=\linewidth]{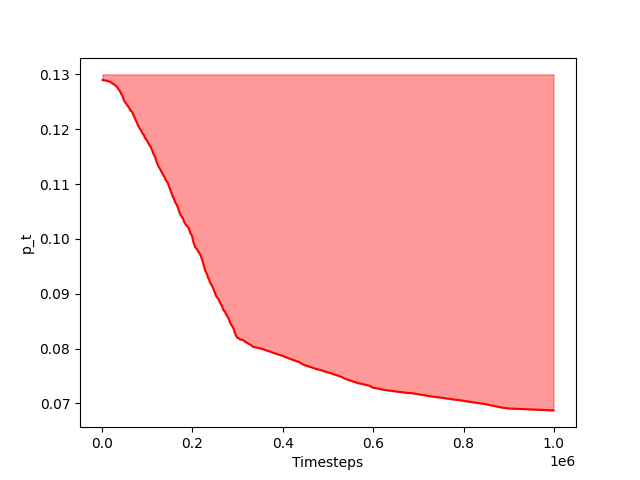}
        \caption*{Reparam-R curriculum $p_i \in (p_t, 0.13)$}
    \end{subfigure}

    \begin{subfigure}{0.48\textwidth}
        \centering
        \includegraphics[width=\linewidth]{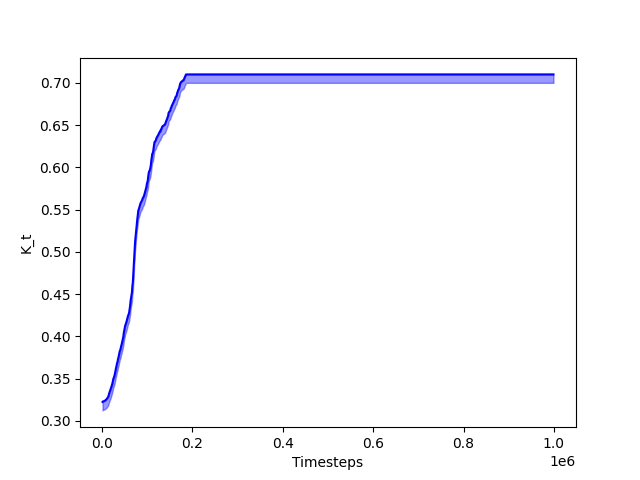}
        \caption*{Frontier curriculum $\kappa_i \in (\kappa_t-0.001, \kappa_t)$}
    \end{subfigure}
    \hfill
    \begin{subfigure}{0.48\textwidth}
        \centering
        \includegraphics[width=\linewidth]{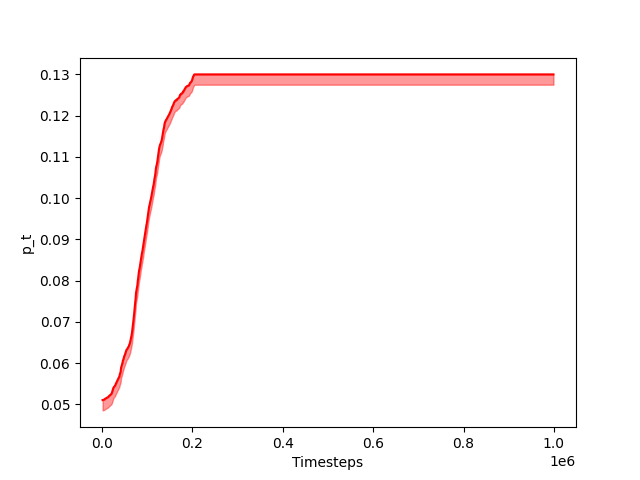}
        \caption*{Frontier curriculum $p_i \in (p_t-0.0001, p_t)$}
    \end{subfigure}
    \caption*{Figure 3: Curriculum Growth Curves and Sampling Ranges- Car Racing}
\end{figure}

\begin{table}[h!]   
\caption{Per-Seed Average Testing Performance across 500 evaluation environments and IQM- Car
Racing.}
\begin{center}
\begin{tabular}{ |c|c|c|c|c|c|c| } 
\hline
\textbf{Model} & \textbf{Seed 0} & \textbf{Seed 1} & \textbf{Seed 2} & \textbf{Seed 3} & \textbf{Seed 4} & \textbf{Average} \\ 
 \hline
 Vanilla & $306 \pm 182$ & $548 \pm 175$ & $608 \pm 189$ & $594 \pm 157$ & $645 \pm 184$ & $540 \pm 178$ \\
 (IQM)  & $322$ & $567$ & $647$ & $615$ & $675$ & $565$ \\
 \hline
  Random & $598 \pm 169$ & $570 \pm 195$ & $592 \pm 217$ & $404 \pm 175$ & $621 \pm 114$ & $557 \pm 174$ \\ 
  (IQM) & $624$ & $595$ & $628$ & $402$ & $622$ & $574$ \\
  \hline
  Manual & $519 \pm 115$ & $702 \pm 183$ & $680 \pm 175$ & $495 \pm 145$ & $688 \pm 212$ & $617 \pm 166$ \\ 
 (IQM) & $536$ & $749$ & $717$ & $516$ & $747$ & $653$ \\
 \hline
 SPRL & $214 \pm 120$ & $564 \pm 130$ & $746 \pm 141$ & $587 \pm 135$ & $530 \pm 139$ & $528 \pm 133$ \\ 
 (IQM) & $210$ & $568$ & $779$ & $602$ & $549$ & $537$ \\
 \hline
 ALP-GMM & $544 \pm 165$ & $114 \pm 98$ & $728 \pm 129$ & $601 \pm 160$ & $553 \pm 188$ & $508 \pm 148$ \\ 
 (IQM) & $581$ & $106$ & $753$ & $623$ & $575$ & $528$ \\
 \hline
  Reparam & $795 \pm 119$ & $613 \pm 186$ & $703 \pm 188$ & $6 \pm 40$ & $126 \pm 81$ & $448 \pm 122$ \\ 
  (IQM) & $829$ & $662$ & $751$ & $3$ & $126$ & $474$ \\
  \hline
  \textbf{Reparam-M} & $531 \pm 143$ & $699 \pm 103$ & $702 \pm 186$ & $607 \pm 144$ & $710 \pm 725$ & $\mathbf{650 \pm 134}$ \\ 
  \textbf{(IQM)} & $547$ & $720$ & $753$ & $622$ & $725$ & $\mathbf{673}$ \\
  \hline
  Reparam-A & $552 \pm 176$ & $628 \pm 155$ & $656 \pm 166$ & $650 \pm 110$ & $628 \pm 161$ & $623 \pm 153$\\
  (IQM) & $588$ & $658$ & $702$ & $663$ & $648$ & $652$ \\
  \hline
  Reparam-R & $199 \pm 123$ & $657 \pm 142$ & $633 \pm 117$ & $648 \pm 204$ & $575 \pm 150$ & $542 \pm 147$ \\ 
  (IQM) & $199$ & $691$ & $648$ & $704$ & $591$ & $567$ \\
  \hline
 Frontier & $358 \pm 168$ & $573 \pm 156$ & $431 \pm 174$ & $569 \pm 118$ & $569 \pm 184$ & $500 \pm 160$\\
 (IQM) & $342$ & $598$ & $438$ & $582$ & $596$ & $511$ \\
 \hline
\end{tabular}
\label{table6}
\end{center}
\end{table}

\clearpage

\section{Results- Bipedal Walker}

Below, the training evaluation curves and curriculum growth curves are presented for all experiments reported in the main paper for the Bipedal Walker environment, with all curves averaged over five random seeds (0,1,2,3,4). Additionally, the interquartile mean (IQM) and per-seed average rewards are reported for each method.

\begin{figure*}[htbp!]
    \centering
    \begin{subfigure}{0.425\linewidth}
        \centering
        \includegraphics[width=\linewidth]{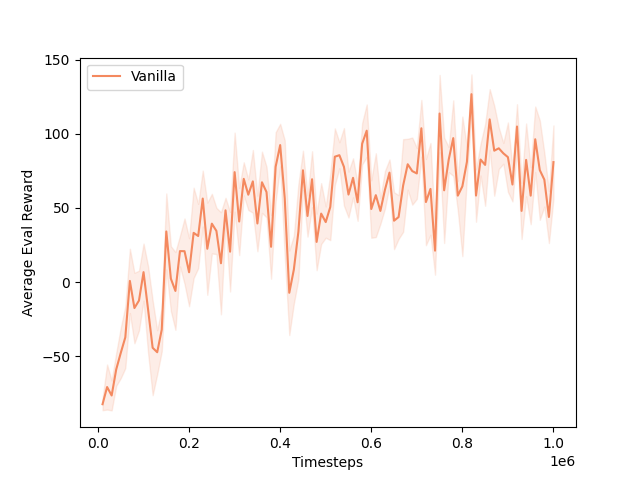}
        \caption*{Vanilla Policy}
        \label{fig1a}
    \end{subfigure}
    \hfill
    \begin{subfigure}{0.45\linewidth}
        \centering
        \includegraphics[width=\linewidth]{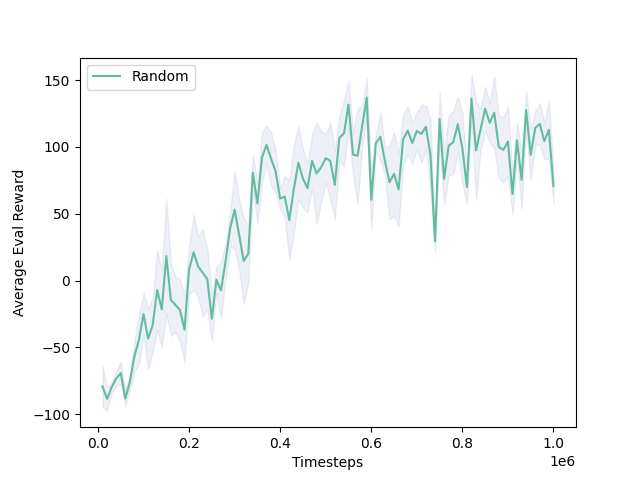}
        \caption*{Random Sampling}
        \label{fig1b}
    \end{subfigure}

    \begin{subfigure}{0.45\linewidth}
        \centering
        \includegraphics[width=\linewidth]{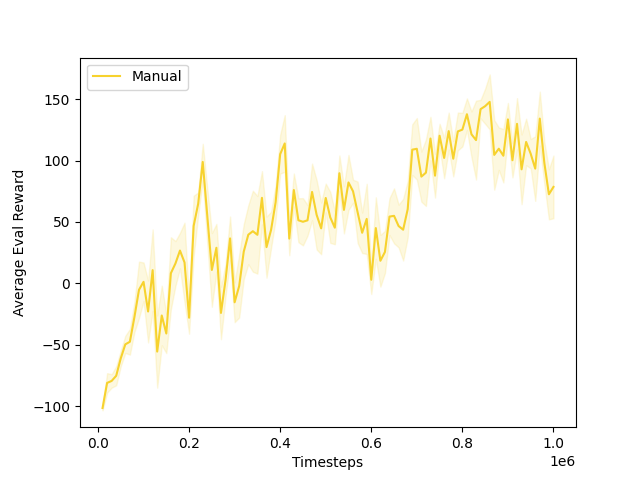}
        \caption*{Manual Curriculum}
        \label{fig2a}
    \end{subfigure}
    \hfill
    \begin{subfigure}{0.45\linewidth}
        \centering
        \includegraphics[width=\linewidth]{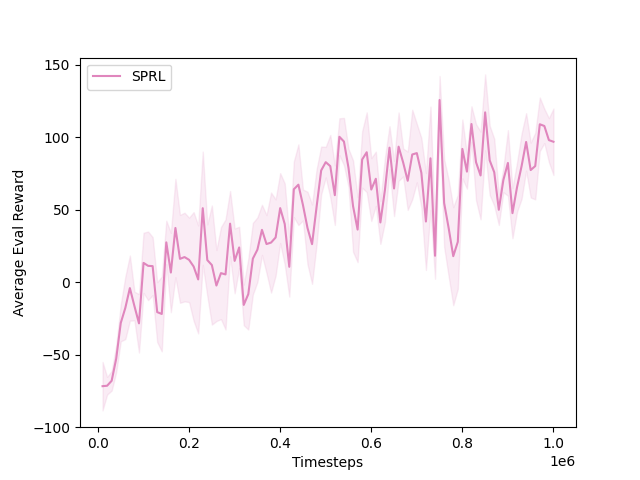}
        \caption*{Self-Paced RL}
        \label{fig2b}
    \end{subfigure}

    \begin{subfigure}{0.45\linewidth}
        \centering
        \includegraphics[width=\linewidth]{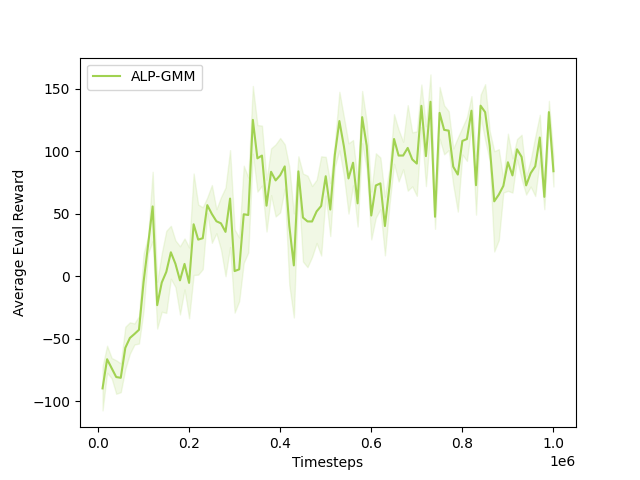}
        \caption*{ALP-GMM}
        \label{fig3a}
    \end{subfigure}
    \hfill
    \begin{subfigure}{0.45\linewidth}
        \centering
        \includegraphics[width=\linewidth]{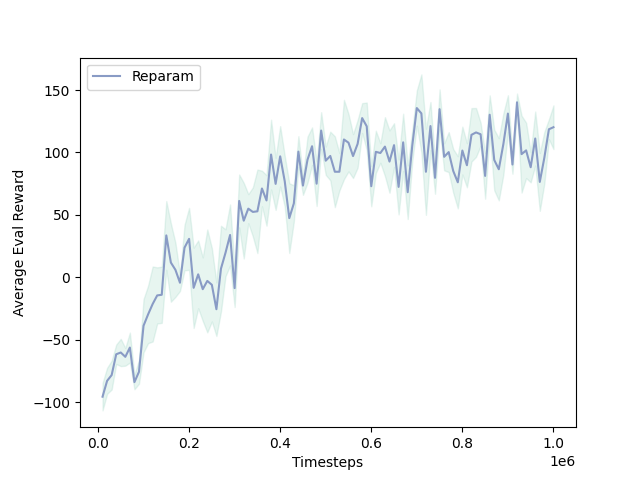}
        \caption*{Reparam}
        \label{fig3b}
    \end{subfigure}
\end{figure*}

\begin{figure*}

    \begin{subfigure}{0.45\linewidth}
        \centering
        \includegraphics[width=\linewidth]{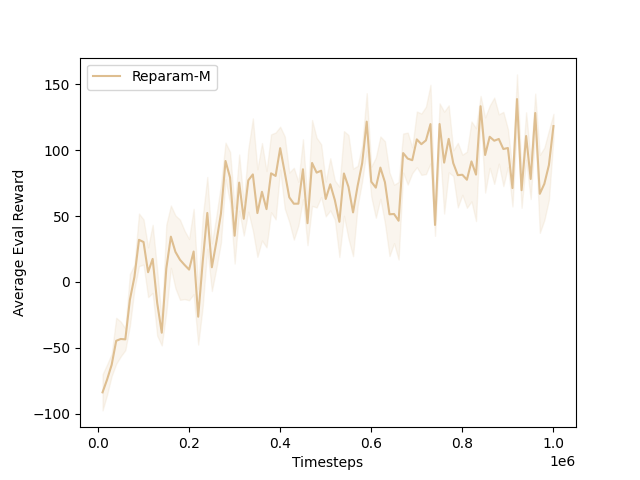}
        \caption*{Reparam-M}
        \label{fig4a}
    \end{subfigure}
    \hfill
    \begin{subfigure}{0.45\linewidth}
        \centering
        \includegraphics[width=\linewidth]{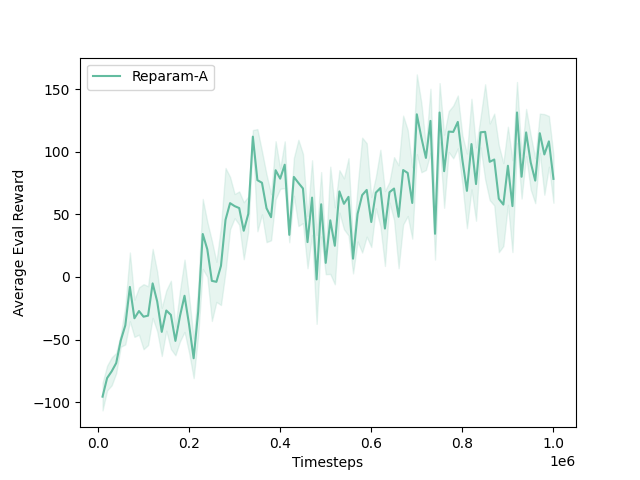}
        \caption*{Reparam-A}
        \label{fig4b}
    \end{subfigure}
    
    \begin{subfigure}{0.45\linewidth}
        \centering
        \includegraphics[width=\linewidth]{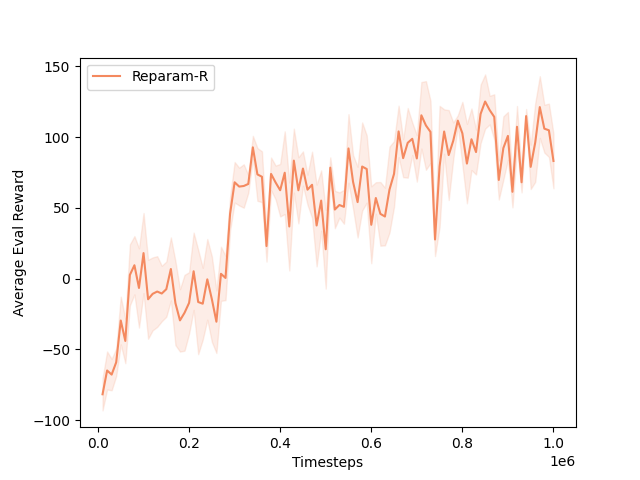}
        \caption*{Reparam-R}
        \label{fig4a}
    \end{subfigure}
    \hfill
    \begin{subfigure}{0.45\linewidth}
        \centering
        \includegraphics[width=\linewidth]{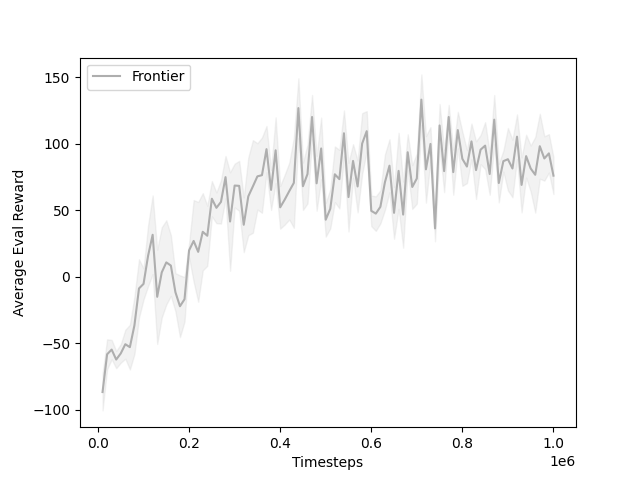}
        \caption*{Frontier}
        \label{fig4b}
    \end{subfigure}
    \caption*{Figure 4: Mean evaluation performance measured every 10000 training timesteps. Curves show the average evaluation reward across 5 random seeds; shaded regions indicate ±1 standard error of the mean. Each evaluation consists of rollouts over 10 environments-Bipedal Walker}
\end{figure*}

\begin{figure*}[htbp!]
    \centering
    \begin{subfigure}{0.3\textwidth}
        \centering
        \includegraphics[width=\linewidth,height=4cm]{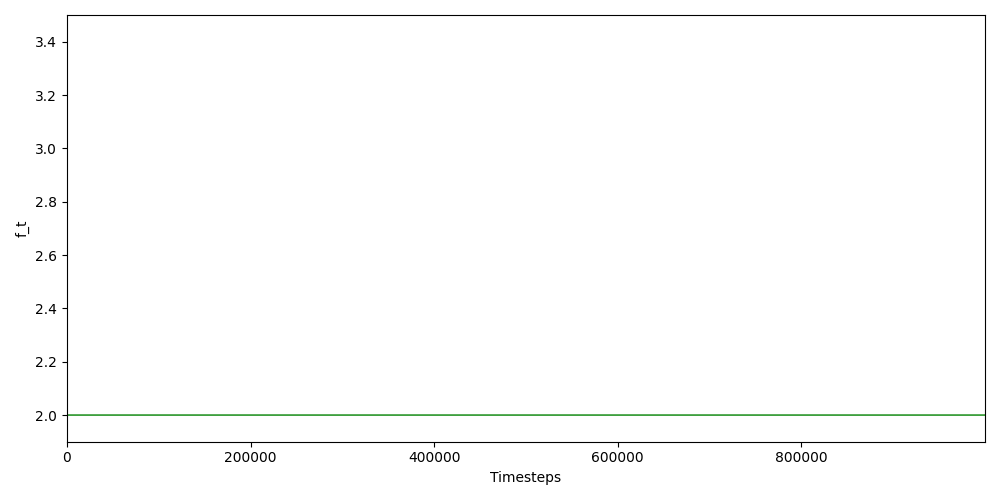}
        \caption*{Vanilla Policy- No Curriculum $f_t=2.5$}
    \end{subfigure}
    \hfill
    \begin{subfigure}{0.3\textwidth}
        \centering
        \includegraphics[width=\linewidth,height=4cm]{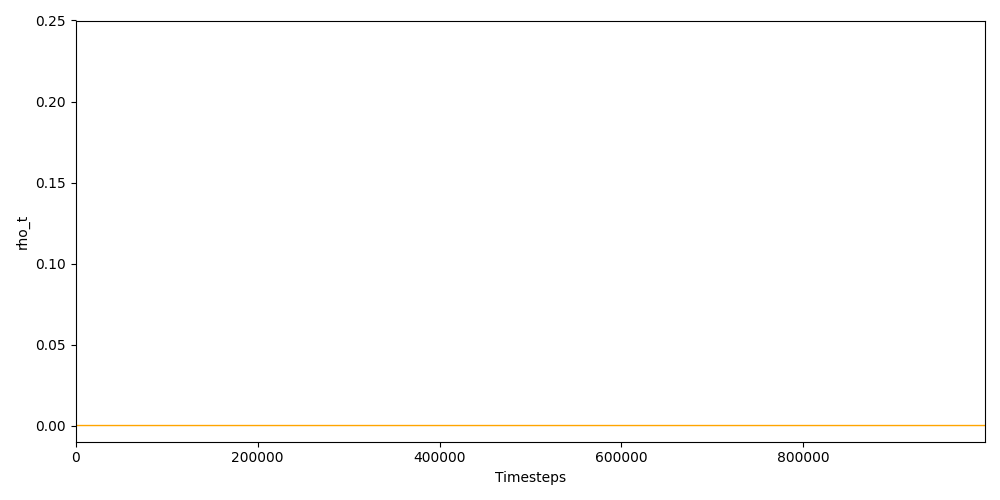}
        \caption*{Vanilla Policy- No Curriculum $\rho_t=0.0$}
    \end{subfigure}
    \hfill
    \begin{subfigure}{0.3\textwidth}
        \centering
        \includegraphics[width=\linewidth,height=4cm]{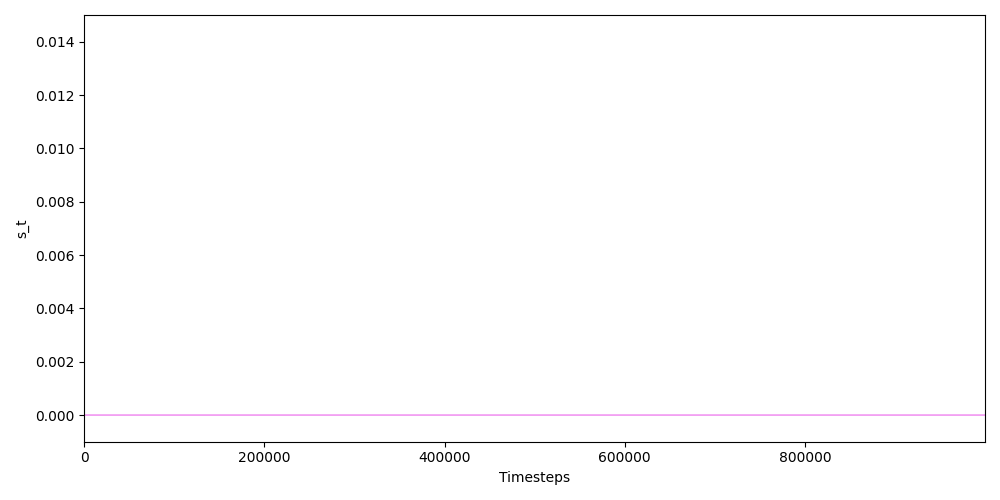}
        \caption*{Vanilla Policy- No Curriculum $s_t=0.0$}
    \end{subfigure}
    
    \begin{subfigure}{0.3\textwidth}
        \centering
        \includegraphics[width=\linewidth,height=4cm]{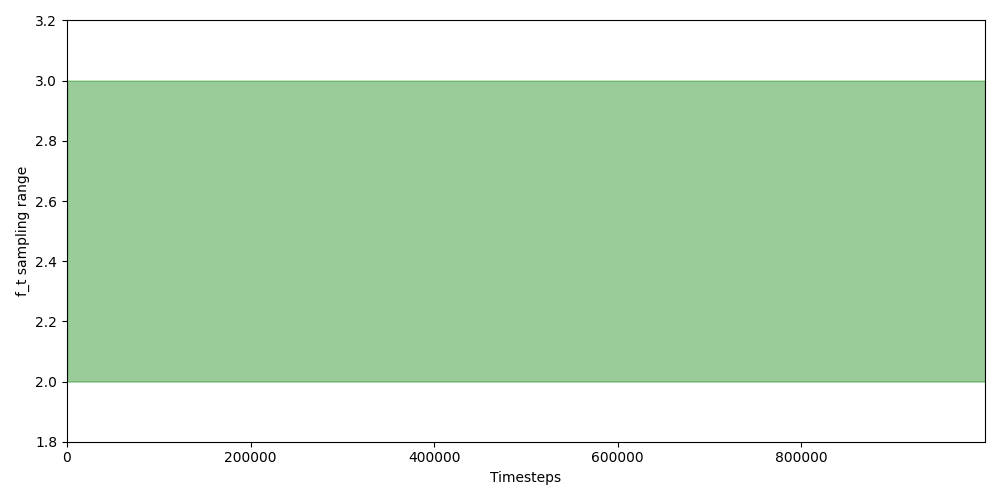}
        \captionsetup{justification=raggedright}
        \caption*{ Random Sampling-  $f_t \in (2.0, 3.0)$}
    \end{subfigure}
    \hfill
    \begin{subfigure}{0.3\textwidth}
        \centering
        \includegraphics[width=\linewidth,height=4cm]{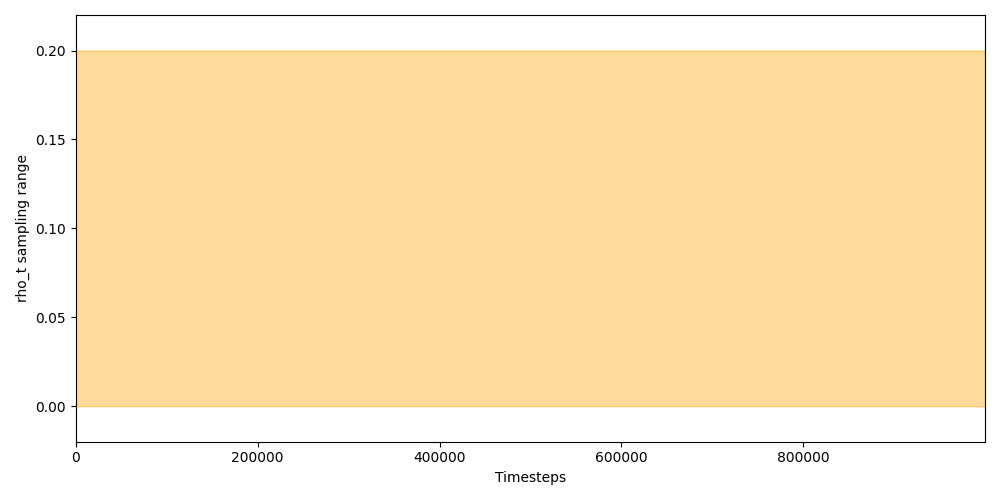}
        \captionsetup{justification=raggedright}
        \caption*{ Random Sampling-  $\rho_t \in (0.0, 0.2)$}
    \end{subfigure}
    \hfill
    \begin{subfigure}{0.3\textwidth}
        \centering
        \includegraphics[width=\linewidth,height=4cm]{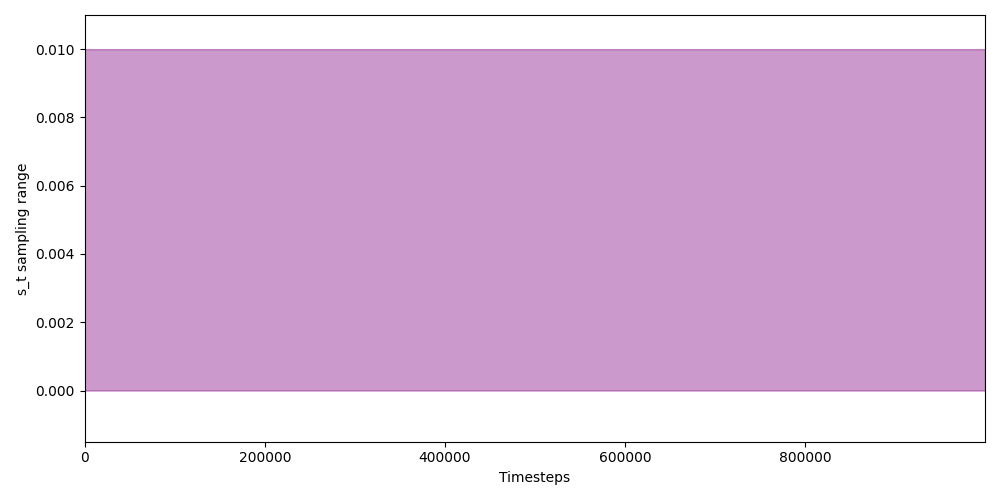}
        \captionsetup{justification=raggedright}
        \caption*{ Random Sampling-  $s_t \in (0.0, 0.01)$}
    \end{subfigure}

\end{figure*}

\begin{figure*}[htbp!]

    \begin{subfigure}{0.32\textwidth}
        \centering
        \includegraphics[width=\linewidth]{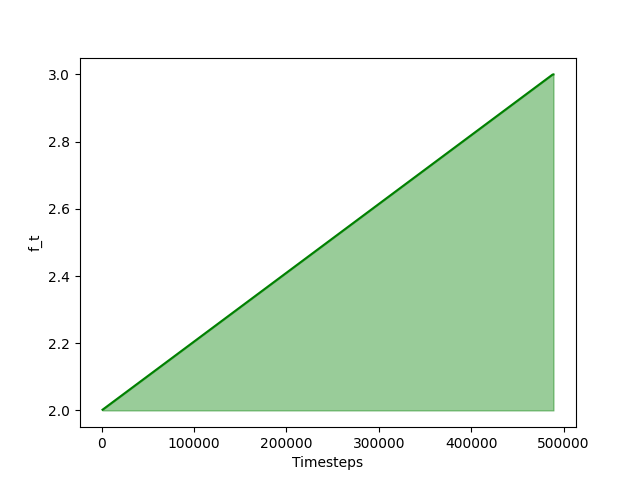}
        \caption*{Manual curriculum $f_i \in (2.0, f_t)$}
    \end{subfigure}
    \hfill
    \begin{subfigure}{0.32\textwidth}
        \centering
        \includegraphics[width=\linewidth]{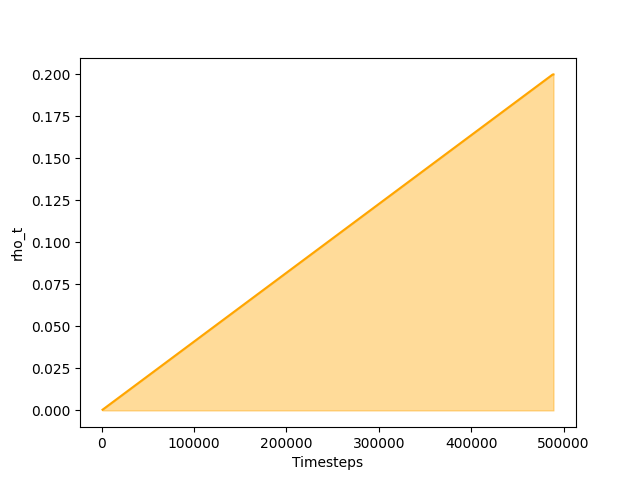}
        \caption*{Manual curriculum $\rho_i \in (0.0, \rho_t)$}
    \end{subfigure}
    \hfill
    \begin{subfigure}{0.32\textwidth}
        \centering
        \includegraphics[width=\linewidth]{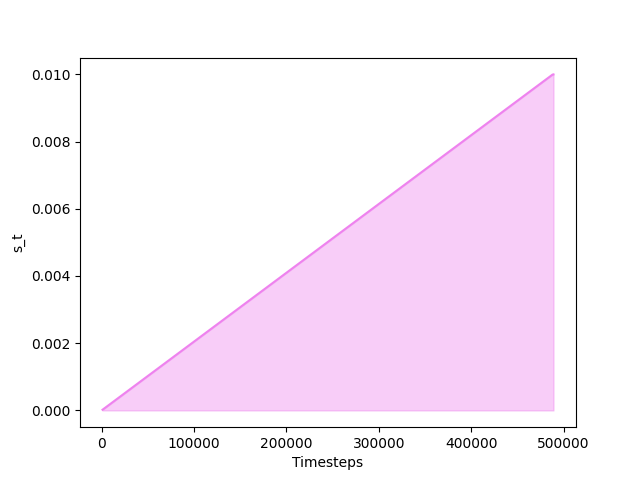}
        \caption*{Manual curriculum $s_i \in (0.0, s_t)$}
    \end{subfigure}

    \begin{subfigure}{0.32\textwidth}
        \centering
        \includegraphics[width=\linewidth]{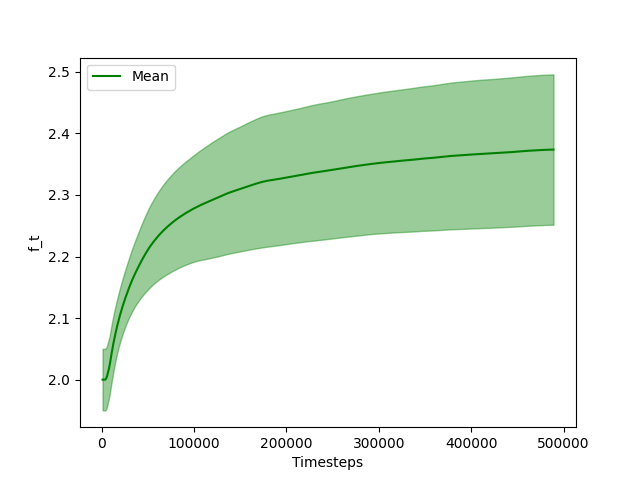}
        \caption*{Self-paced RL $f_i \sim \mathcal{N}(\mu_t^f,\Sigma_t^f)$}
    \end{subfigure}
    \hfill
    \begin{subfigure}{0.32\textwidth}
        \centering
        \includegraphics[width=\linewidth]{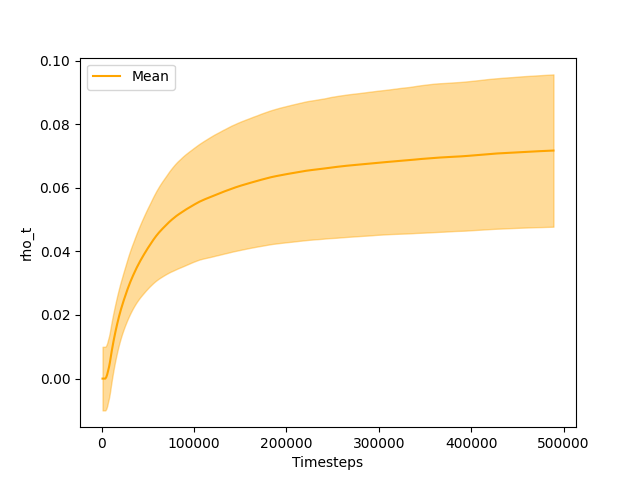}
        \caption*{Self-paced RL $\rho_i \sim \mathcal{N}(\mu_t^\rho,\Sigma_t^\rho)$}
    \end{subfigure}
    \hfill
    \begin{subfigure}{0.32\textwidth}
        \centering
        \includegraphics[width=\linewidth]{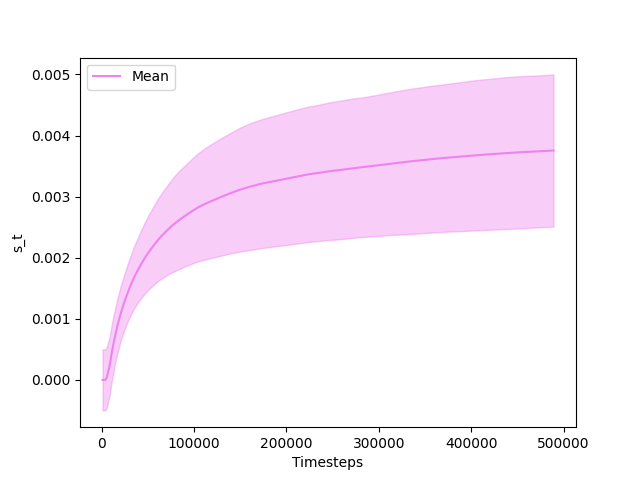}
        \caption*{Self-paced RL $s_i \sim \mathcal{N}(\mu_t^s,\Sigma_t^s)$}
    \end{subfigure}
    
    \begin{subfigure}{0.32\textwidth}
        \centering
        \includegraphics[width=\linewidth]{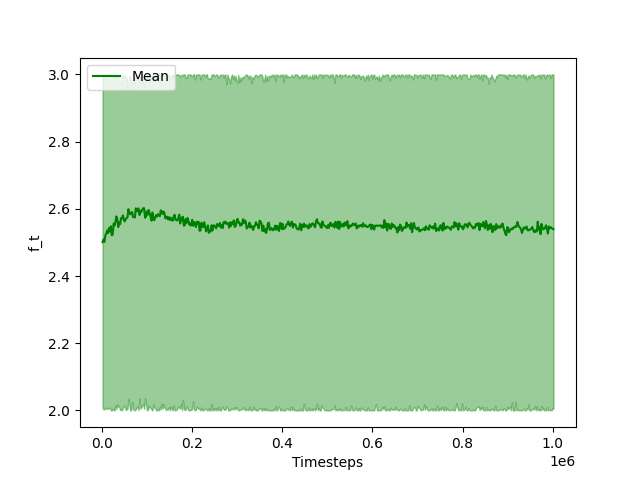}
        \captionsetup{justification=raggedright}
        \caption*{ALP-GMM $f_i \sim \sum_{k=1}^{K} w_k \mathcal{N}(\mu_k^f,\Sigma_k^f)$}
    \end{subfigure}
    \hfill
    \begin{subfigure}{0.32\textwidth}
        \centering
        \includegraphics[width=\linewidth]{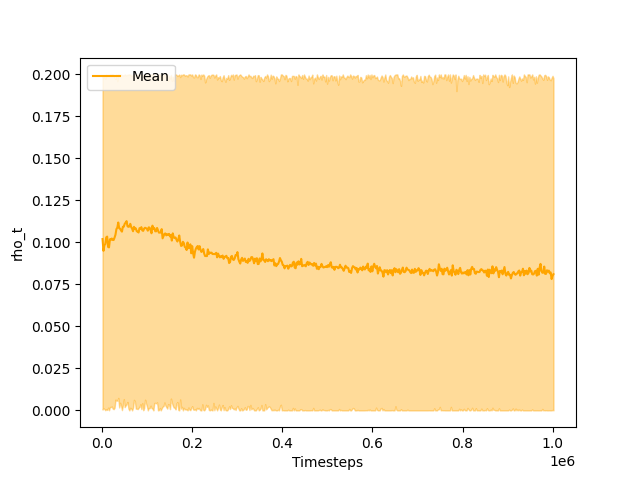}
        \captionsetup{justification=raggedright}
        \caption*{ALP-GMM $\rho_i \sim \sum_{k=1}^{K} w_k \mathcal{N}(\mu_k^\rho,\Sigma_k^\rho)$}
    \end{subfigure}
    \hfill
    \begin{subfigure}{0.32\textwidth}
        \centering
        \includegraphics[width=\linewidth]{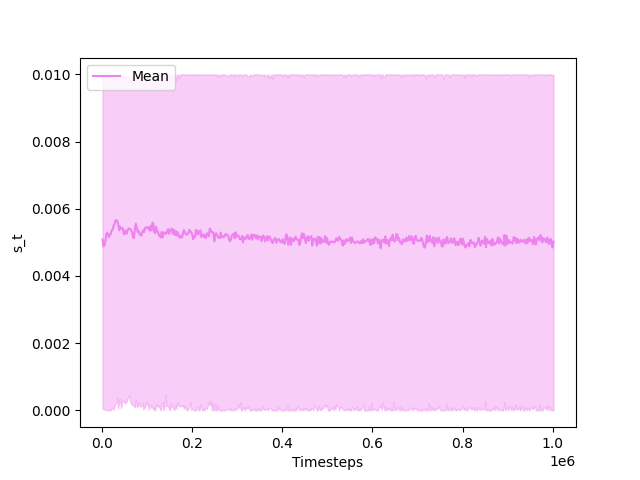}
        \captionsetup{justification=raggedright}
        \caption*{ALP-GMM $s_i \sim \sum_{k=1}^{K} w_k \mathcal{N}(\mu_k^s,\Sigma_k^s)$}
    \end{subfigure}

    \begin{subfigure}{0.32\textwidth}
        \centering
        \includegraphics[width=\linewidth]{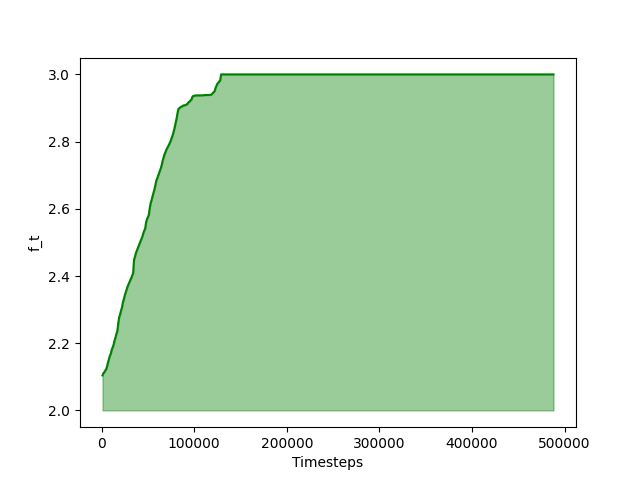}
        \captionsetup{justification=raggedright}
        \caption*{Reparam curriculum $f_i \in (2.0, f_t)$}
    \end{subfigure}
    \hfill
    \begin{subfigure}{0.32\textwidth}
        \centering
        \includegraphics[width=\linewidth]{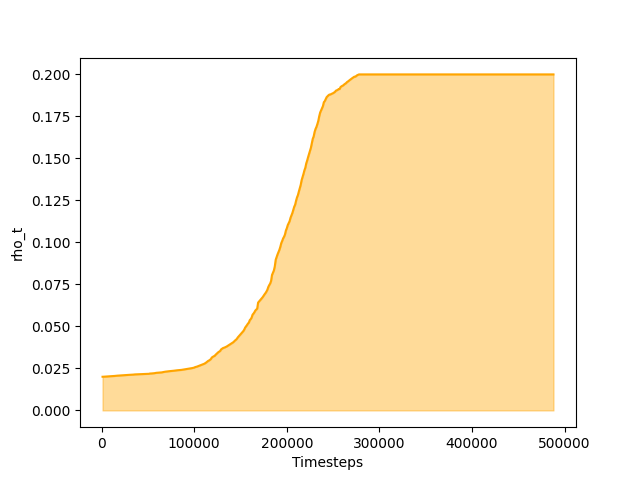}
        \captionsetup{justification=raggedright}
        \caption*{Reparam curriculum $\rho_i \in (0.0, \rho_t)$}
    \end{subfigure}
    \hfill
    \begin{subfigure}{0.32\textwidth}
        \centering
        \includegraphics[width=\linewidth]{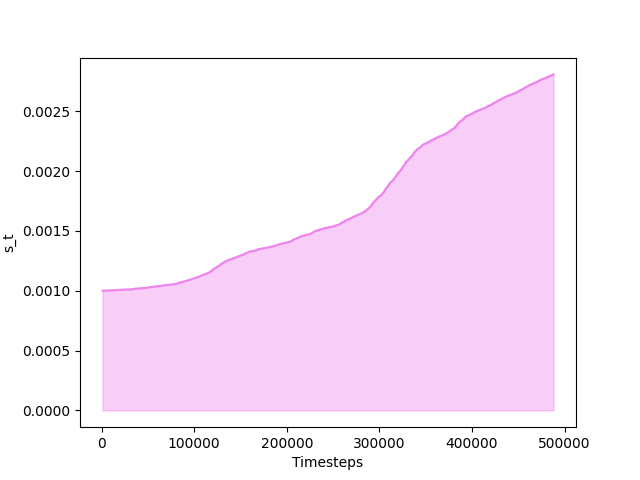}
        \captionsetup{justification=raggedright}
        \caption*{Reparam curriculum $s_i \in (0.0, s_t)$}
    \end{subfigure}
    
    \begin{subfigure}{0.32\textwidth}
        \centering
        \includegraphics[width=\linewidth]{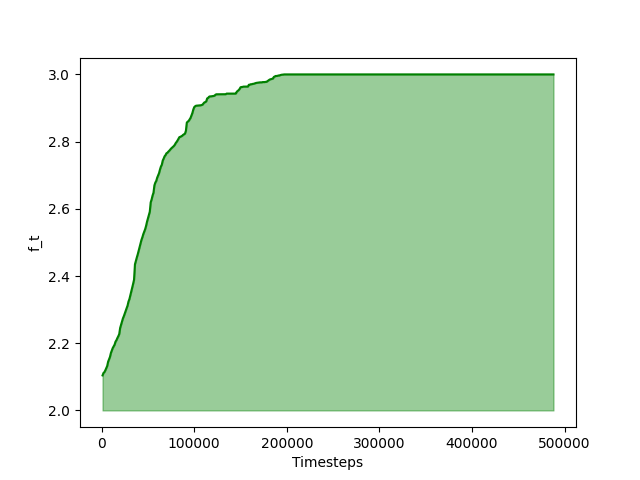}
        \captionsetup{justification=raggedright}
        \caption*{Reparam-M curriculum $f_i \in (2.0, f_t)$}
    \end{subfigure}
    \hfill
    \begin{subfigure}{0.32\textwidth}
        \centering
        \includegraphics[width=\linewidth]{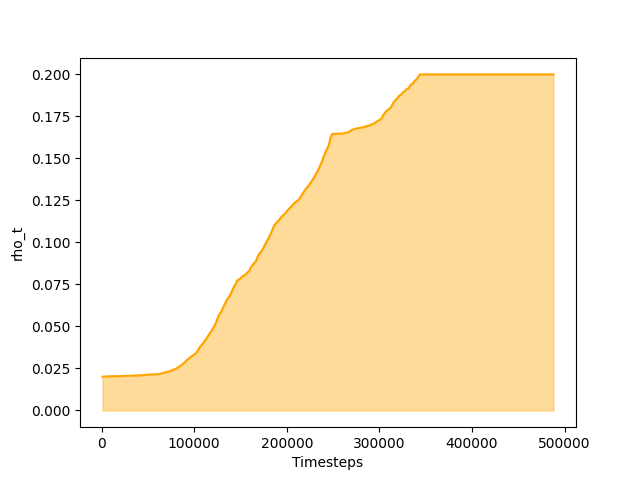}
        \captionsetup{justification=raggedright}
        \caption*{Reparam-M curriculum $\rho_i \in (0.0, \rho_t)$}
    \end{subfigure}
    \hfill
    \begin{subfigure}{0.32\textwidth}
        \centering
        \includegraphics[width=\linewidth]{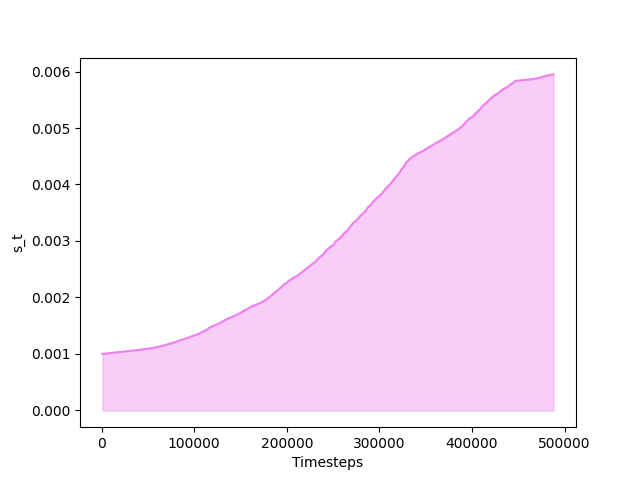}
        \captionsetup{justification=raggedright}
        \caption*{Reparam-M curriculum $s_i \in (0.0, s_t)$}
    \end{subfigure}

\end{figure*}

\begin{table}[h!]   
\caption{Per-seed Average and IQM of Testing Rewards- Bipedal Walker}
\begin{center}
\begin{tabular}{ |c|c|c|c|c|c|c| } 
\hline
Model & Seed 0 & Seed 1 & Seed 2 & Seed 3 & Seed 4 & Average \\ 
 \hline
 Vanilla & $111 \pm 140$ & $102 \pm 150$ & $59 \pm 139$ & $106 \pm 147$ & $70 \pm 145$ & $90 \pm 144$ \\ 
 (IQM) & $111$ & $92$ & $17$ & $91$ & $36$ & $69$ \\
 \hline
  Random & $83 \pm 120$ & $95 \pm 140$ & $125 \pm 121$ & $115 \pm 136$ & $138 \pm 145$ & $111 \pm 132$ \\ 
  (IQM) & $86$ & $94$ & $157$ & $128$ & $158$ & $125$ \\
  \hline
  Manual & $134 \pm 139$ & $127 \pm 140$ & $165 \pm 134$ & $37 \pm 156$ & $121 \pm 139$ & $117 \pm 142$ \\ 
  (IQM) & $151$ & $145$ & $202$ & $-7$ & $136$ & $125$ \\
 \hline
 SPRL & $58 \pm 147$ & $131 \pm 142$ & $70 \pm 129$ & $82 \pm 143$ & $101 \pm 147$ & $88 \pm 142$ \\ 
 (IQM) & $18$ & $149$ & $47$ & $49$ & $108$ & $74$ \\
 \hline
 ALP-GMM & $127 \pm 143$ & $103 \pm 138$ & $97 \pm 139$ & $143 \pm 141$ & $109 \pm 137$ & $116 \pm 139$ \\ 
 (IQM) & $135$ & $109$ & $94$ & $168$ & $121$ & $125$ \\
 \hline
  \textbf{Reparam} & $124 \pm 126$ & $121 \pm 149$ & $101 \pm 142$ & $136 \pm 126$ & $152 \pm 136$ & $\mathbf{127 \pm 136}$ \\ 
  \textbf{(IQM)} & $141$ & $130$ & $94$ & $156$ & $156$ & $\mathbf{141}$ \\
  \hline
  Reparam-M & $140 \pm 141$ & $64 \pm 131$ & $144 \pm 136$ & $70 \pm 109$ & $144 \pm 136$ & $112 \pm 131$\\ 
  (IQM) & $157$ & $42$ & $173$ & $69$ & $167$ & $122$ \\
  \hline
  Reparam-A & $143 \pm 140$ & $61 \pm 125$ & $142 \pm 132$ & $96 \pm 143$ & $106 \pm 145$ & $110 \pm 137$ \\ 
  (IQM) & $172$ & $37$ & $167$ & $83$ & $102$ & $112$ \\
  \hline
  Reparam-R & $108 \pm 122$ & $82 \pm 127$ & $127 \pm 140$ & $136 \pm 128$ & $91 \pm 123$ & $109 \pm 128$\\ 
  (IQM) & $120$ & $72$ & $149$ & $165$ & $101$ & $121$ \\
  \hline
 Frontier & $95 \pm 143$ & $74 \pm 137$ & $109 \pm 147$ & $100 \pm 127$ & $87 \pm 139$ & $93 \pm 139$\\ 
 (IQM) & $82$ & $50$ & $106$ & $93$ & $78$ & $82$ \\
 \hline
\end{tabular}
\label{table5}
\end{center}
\end{table}

\begin{figure*}[htbp!]
    \begin{subfigure}{0.32\textwidth}
        \centering
        \includegraphics[width=\linewidth]{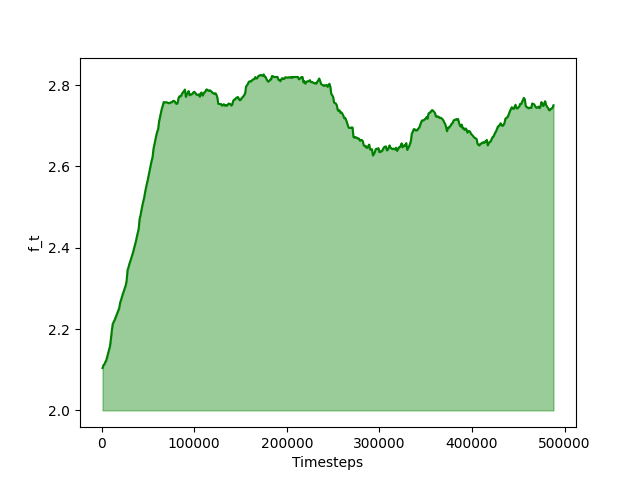}
        \captionsetup{justification=raggedright}
        \caption*{Reparam-A curriculum $f_i \in (2.0, f_t)$}
    \end{subfigure}
    \hfill
    \begin{subfigure}{0.32\textwidth}
        \centering
        \includegraphics[width=\linewidth]{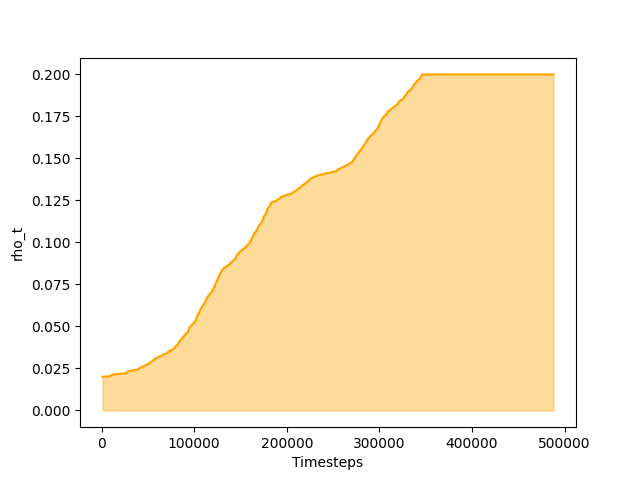}
        \captionsetup{justification=raggedright}
        \caption*{Reparam-A curriculum $\rho_i \in (0.0, \rho_t)$}
    \end{subfigure}
    \hfill
    \begin{subfigure}{0.32\textwidth}
        \centering
        \includegraphics[width=\linewidth]
        {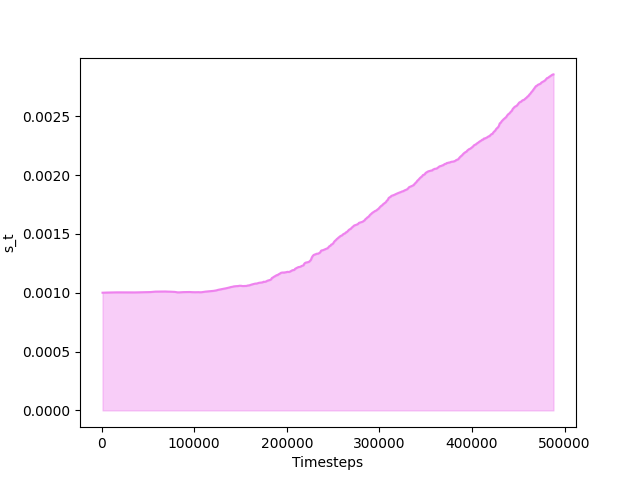}
        \captionsetup{justification=raggedright}
        \caption*{Reparam-A curriculum $s_i \in (0.0, s_t)$}
    \end{subfigure}
    
    \begin{subfigure}{0.32\textwidth}
        \centering
        \includegraphics[width=\linewidth]{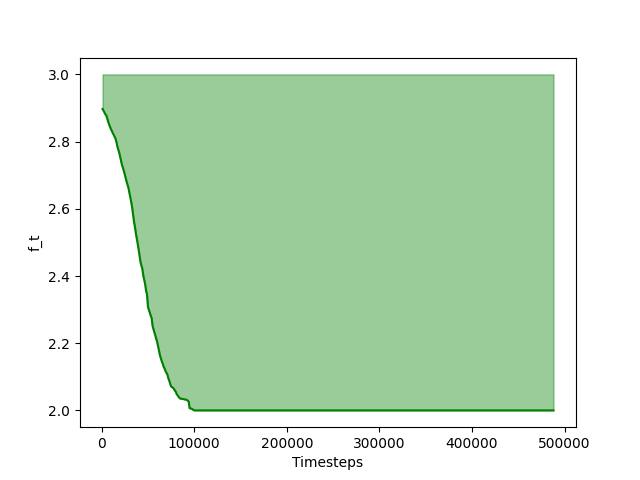}
        \captionsetup{justification=raggedright}
        \caption*{Reparam-R curriculum $f_i \in (f_t, 3.0)$}
    \end{subfigure}
    \hfill
    \begin{subfigure}{0.32\textwidth}
        \centering
        \includegraphics[width=\linewidth]{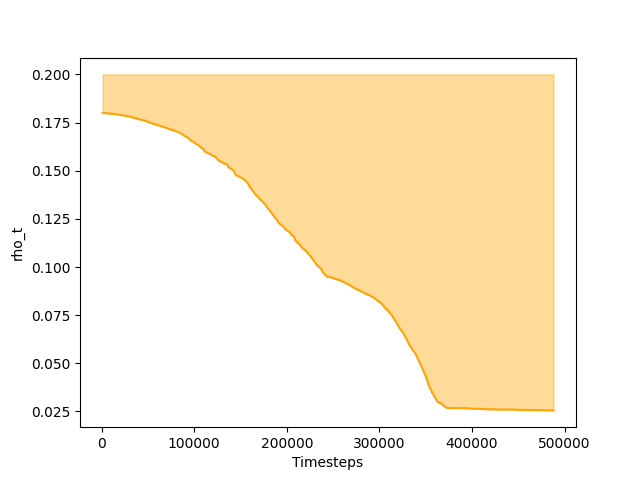}
        \captionsetup{justification=raggedright}
        \caption*{Reparam-R curriculum $\rho_i \in (\rho_t, 0.2)$}
    \end{subfigure}
    \hfill
    \begin{subfigure}{0.32\textwidth}
        \centering
        \includegraphics[width=\linewidth]{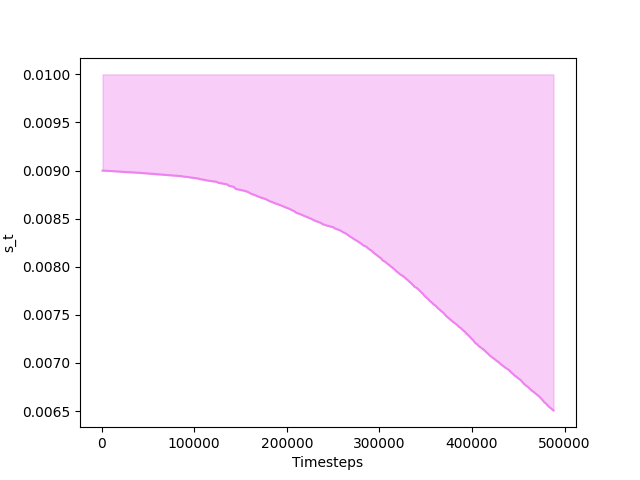}
        \captionsetup{justification=raggedright}
        \caption*{Reparam-R curriculum $s_i \in (s_t, 0.01)$}
    \end{subfigure}

\end{figure*}

\clearpage
\begin{figure}[!htbp]
    \begin{subfigure}{0.32\textwidth}
        \centering
        \includegraphics[width=\linewidth]{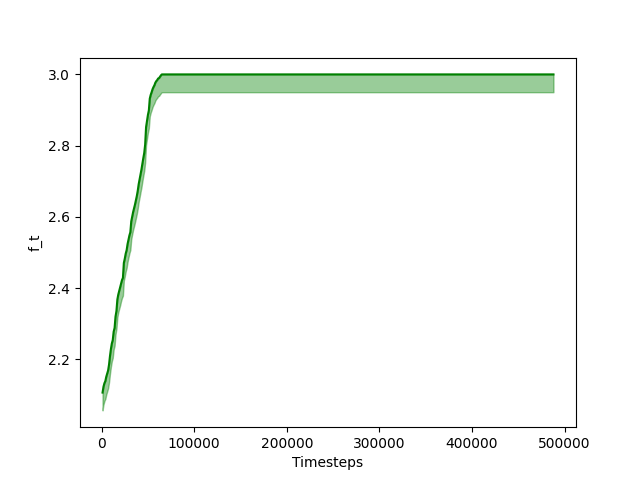}
        \captionsetup{justification=raggedright}
        \caption*{Frontier curriculum $f_i \in (f_t-0.001, f_t)$}
    \end{subfigure}
    \hfill
    \begin{subfigure}{0.32\textwidth}
        \centering
        \includegraphics[width=\linewidth]{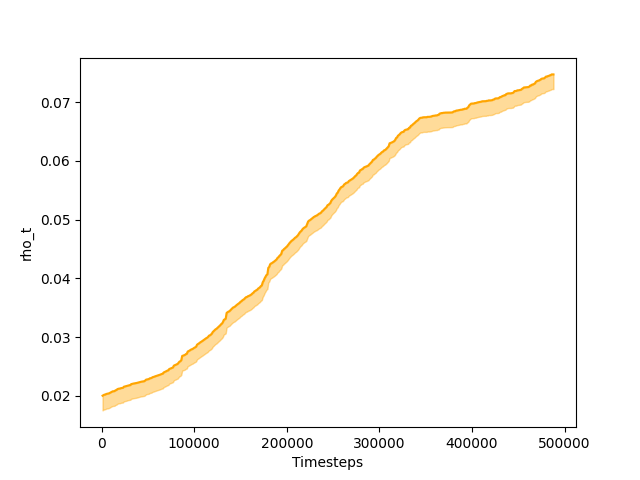}
        \captionsetup{justification=raggedright}
        \caption*{Frontier curriculum $\rho_i \in (\rho_t-0.0001, \rho_t)$}
    \end{subfigure}
    \hfill
    \begin{subfigure}{0.32\textwidth}
        \centering
        \includegraphics[width=\linewidth]{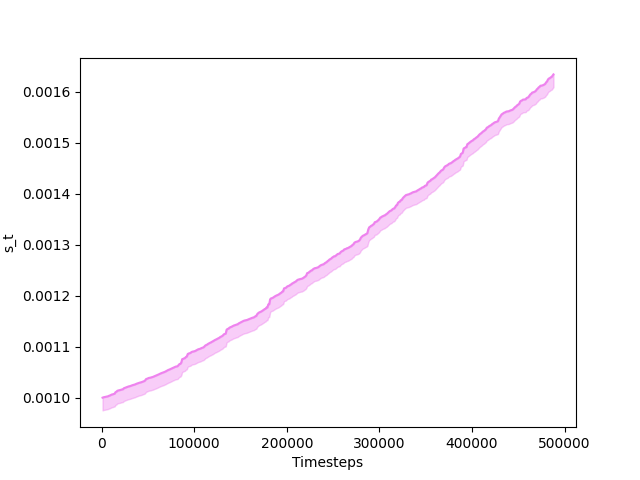}
        \captionsetup{justification=raggedright}
        \caption*{Frontier curriculum $s_i \in (s_t-0.0001, s_t)$}
    \end{subfigure}
    \caption*{Figure 5: Curriculum Growth Curves and Sampling Ranges- Bipedal Walker}
\end{figure}

Renderings of all the policies trained with the curricula presented in the main paper are included in this
\href{https://github.com/PRISHIta123/adversarial_curriculum_for_navigation_policies}{GitHub repository}.

\vspace*{\fill}
\clearpage

\end{document}